\pdfoutput=1
\documentclass[preprint,12pt]{elsarticle}

\usepackage{amssymb}
\usepackage{amsmath}
\usepackage{url}
\usepackage{xcolor}
\colorlet{red}{black}
\newcommand{\rev}[1]{\textcolor{red}{#1}}
\newenvironment{redblock}{\begingroup\color{red}}{\endgroup}

\usepackage{graphicx}
\usepackage{array}
\usepackage{multirow}
\usepackage{booktabs}
\usepackage{tabularx}
\usepackage{makecell}

\usepackage{subcaption}
\usepackage{enumitem}
\setlist[itemize]{itemsep=0pt, parsep=0pt, topsep=0pt, partopsep=0pt, leftmargin=*}

\usepackage{float}
\usepackage{placeins}

\usepackage{listings}
\usepackage{pifont}

\usepackage{hyperref}
\usepackage{cleveref}
\usepackage{algorithm}
\usepackage{algpseudocode}
\usepackage{wrapfig}

\makeatletter
\@ifundefined{highlights}{%
  \newenvironment{highlights}{\begin{itemize}}{\end{itemize}}%
}{}
\makeatother

\journal{Neurocomputing}

\begin{document}

\begin{frontmatter}



\title{Beyond Sparse Supervision:
Diffusion-Guided Learning for Few-Shot Graph Fraud Detection}


\author[1]{Liming Liu}
\author[1]{Chao Hu}
\author[2]{Mingfei Lu}
\author[1]{Yiwei Ge}
\author[1]{Xingle Li}
\author[1]{Heyuan Shi}

\address[1]{Central South University, Changsha, China}
\address[2]{University of Technology Sydney, Australian Artificial Intelligence Institute, Sydney, Australia}

%% Abstract
\begin{abstract}
Graph-based fraud detection is \rev{essential} for safeguarding large-scale transaction systems, where undetected anomalies may lead to substantial financial \rev{losses} and security risks.
Real-world fraud graphs pose two coupled challenges: \rev{sparse and imbalanced supervision}, where verified fraudulent labels are scarce and heavily skewed toward benign accounts, and \rev{representation dilution}, where spatial message passing may oversmooth camouflaged anomalies while spectral filters may suppress fraud-relevant mid- and high-frequency irregularities.
To address these challenges, we propose ADC-GNN, \rev{short for Attention-guided Diffusion-Contrastive Graph Neural Network}, a unified framework that combines \rev{diffusion-guided feature augmentation, contrastive representation learning, and multi-hop spectral attention} for few-shot graph fraud detection.
The diffusion component is formulated as a \rev{feature-space denoising augmentation mechanism rather than a full topology-generative graph diffusion model}: it constructs noise-perturbed node-feature views under a cosine schedule and uses contrastive learning to stabilize node representations across perturbations.
The spectral attention module further adaptively emphasizes \rev{fraud-relevant hop-level and relation-level cues}.
We evaluate ADC-GNN primarily on three public benchmarks and additionally report a proprietary real-world telecom transaction dataset with approximately 60{,}000 records \rev{as a private case study}.
Under the 1\% training setting, ADC-GNN achieves consistent improvements over original graph fraud baselines and four protocol-consistent recent graph anomaly/fraud baselines on the public benchmarks.
\rev{Additional analyses on split stability, training ratios, oversampling alternatives, module-level ablations, diffusion schedules, and runtime and memory-consumption comparisons further characterize the effective operating regime of ADC-GNN.}
Our code is released at \url{https://github.com/llmllmllm/ADC-GNN}.
\end{abstract}


\begin{highlights}
\item \rev{Detects fraud graphs with scarce and imbalanced labels.}
\item \rev{Uses diffusion as feature-space denoising augmentation.}
\item \rev{Combines contrastive learning with spectral attention.}
\item \rev{Adds split stability, ratio-wise comparison, oversampling, ablation, sensitivity, and efficiency analyses.}
\end{highlights}

%% Keywords

\end{frontmatter}




\section{Introduction}
With the widespread digitization of financial services, online transaction networks have become increasingly complex, enabling sophisticated and large-scale fraudulent activities. 
Accurately identifying fraudulent accounts in such environments is essential for maintaining system integrity and financial security~\cite{ma2021comprehensive, qiao2024ggad}. 
Graph Neural Networks (GNNs)~\cite{hamilton2017inductive, xu2018powerful, velivckovic2017graph, xu2024revisiting, fang2024graphfa}, including both spatial message passing and spectral filtering variants, have demonstrated strong potential in modeling relational dependencies and capturing structural patterns for fraud detection~\cite{kipf2016semi, chien2020adaptive, tang2023gadbench, li2024graph}, often outperforming traditional feature-based detectors by leveraging relational signals.

\begin{wrapfigure}{r}{0.48\columnwidth}
    \centering
    \vspace{-6pt}
    \includegraphics[width=0.46\columnwidth]{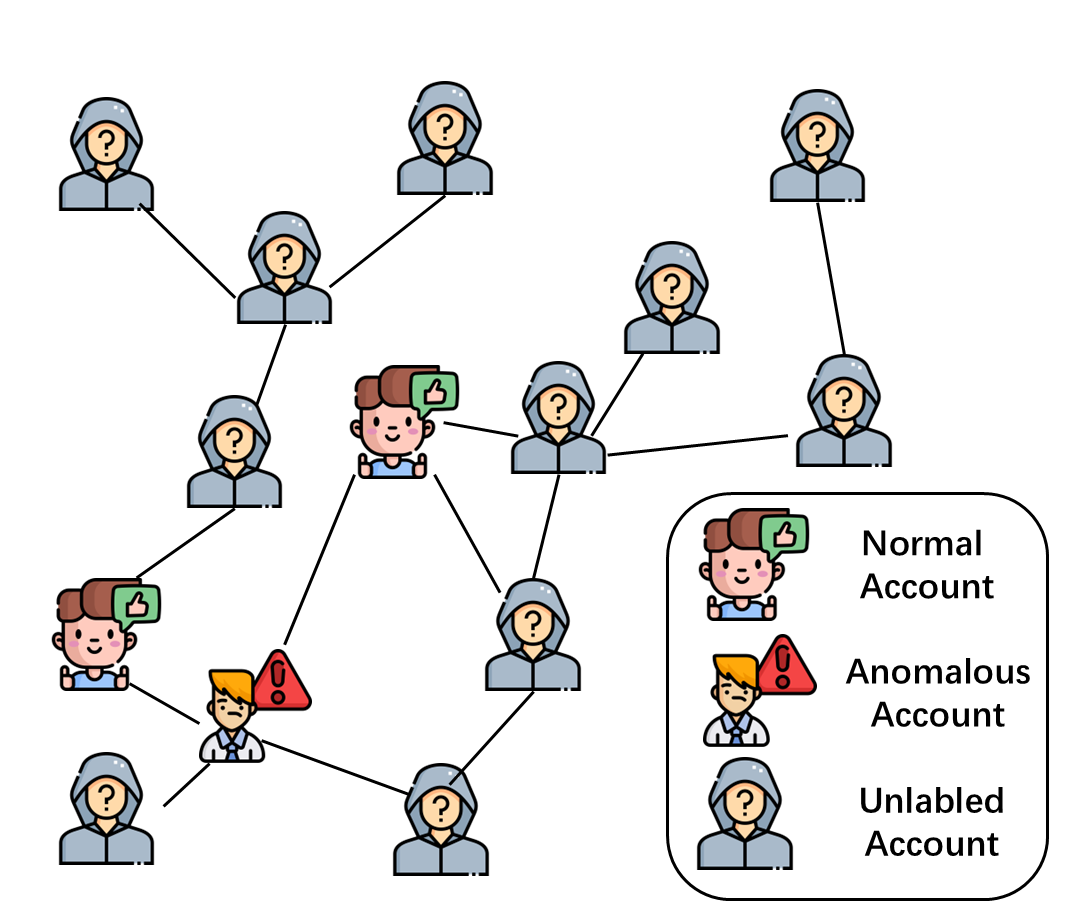}
    \vspace{-6pt}
    \caption{Illustration of the few-shot and label-imbalance challenges in real-world transaction networks.
    Few-shot: only a small portion of nodes are labeled, resulting in insufficient supervision.
    Label imbalance: fraudulent nodes are often buried within clusters of normal nodes.}
    \label{fig_s1_illustration}
\end{wrapfigure}

\textbf{(i) Spatial GNNs} improve robustness via sampling, pruning, or relation-aware aggregation, yet they still rely on message passing that is vulnerable to adversarial neighborhood construction: fraudsters deliberately attach to benign nodes to camouflage their behavior, making local aggregation propagate misleading context and wash out rare anomaly cues.
\textbf{(ii) Spectral GNNs} incorporate frequency-domain filtering to capture global structures, but many filters are biased toward low-frequency smoothness; as a consequence, irregular mid/high-frequency components---often associated with abrupt behavioral deviations and structural inconsistency---can be suppressed, especially when supervision is scarce and gradients are dominated by the benign majority.
\rev{\textbf{(iii) Heterogeneous models} introduce relation-specific modeling for multi-type interactions, but practical deployments may still suffer from semantic collapse when different relation types are merged into a shared representation or when relation-specific fraud semantics are weakly supervised.}

However, two gaps remain under real-world supervision regimes.
\textbf{(1) supervision regime mismatch:} transaction graphs provide sparse and unreliable supervision with extreme class imbalance. Labels are expensive, delayed, and sometimes uncertain, leaving most nodes unlabeled; meanwhile, fraudulent accounts often constitute only a tiny fraction of the population. Under this regime, gradient-based training is biased toward the benign majority, and self-training can amplify majority patterns, causing feature collapse where anomalous embeddings drift toward benign ones. This effect is further exacerbated by intentional camouflage, which creates unreliable propagation paths and injects noise into neighborhood signals.
\textbf{(2) anomaly representation dilution across views:} under weak supervision, existing designs struggle to preserve discriminative anomaly cues consistently across spatial and spectral perspectives. Spatial aggregation tends to oversmooth subtle and long-range anomaly patterns, while spectral smoothing suppresses fraud-relevant high-frequency irregularities. Moreover, heterogeneous relation semantics are not exploited in a complementary spatial--spectral manner: when multi-type interactions are collapsed, relation-specific cues that indicate collusion can be blurred into background noise, further weakening anomaly separability in the learned embedding space.

These gaps jointly motivate the need for a framework that (a) enriches minority supervision without additional labels and (b) preserves discriminative anomaly cues across spatial, spectral, and relational views. 
\begin{redblock}
Accordingly, we propose ADC-GNN, short for \emph{Attention-guided Diffusion-Contrastive Graph Neural Network}, for fraud detection under sparse, imbalanced, and heterogeneous graph settings.
To avoid overstating the role of the diffusion component, we clarify that ADC-GNN does not perform full graph generation with a learned reverse process over both node attributes and edges.
Instead, it uses diffusion-guided feature-space augmentation: node attributes are perturbed under a cosine schedule to construct multiple noisy views, and contrastive learning is used to preserve node identity and stabilize representations across perturbations.
This design provides controlled feature-space regularization without requiring additional manual annotations while avoiding invalid edge perturbations in transaction graphs, where relation semantics are often constrained by real business events.
To enhance structural discrimination, we further design a relation-aware multi-hop spectral attention mechanism that adaptively emphasizes fraud-relevant hop-level and relation-level cues.
The resulting graph--spectral representation combines local topology, relation semantics, and frequency-domain irregularities under a unified classification objective.
\end{redblock}

We evaluate ADC-GNN on three public benchmarks---Amazon~\cite{rayana2015collective}, YelpChi~\cite{mcauley2013amateurs}, 
and T-Finance~\cite{tang2022rethinking}---as well as a proprietary real-world telecom transaction dataset containing approximately 60{,}000 records. 
\begin{redblock}
The public datasets serve as the main reproducibility anchors, while the proprietary CM dataset is reported as a private case-study setting with schema-level disclosure and aggregate statistics.
Experimental results show that ADC-GNN obtains consistent gains at the 1\% training ratio.
We retain the original graph fraud baselines and further include protocol-consistent recent graph anomaly/fraud baselines, including CGAD, ARC, UniGAD, and CGNN.
Methods whose official protocols rely on incompatible training ratios, evaluation metrics, or unreleased preprocessing settings are discussed in related work but are not mixed into the main comparison table, so that the experimental protocol remains internally consistent.
Additional experiments report training-ratio comparisons, oversampling-based baselines, granular ablations, diffusion-schedule comparisons, split stability analysis, and runtime and memory-consumption comparisons to better delimit the effective operating regime and computational cost of ADC-GNN.
\end{redblock}

Our main contributions are summarized as follows:
\begin{itemize}[itemsep=0pt, parsep=0pt, topsep=2pt]
    \item \rev{We formulate few-shot graph fraud detection as a sparse-supervision and representation-dilution problem, and propose ADC-GNN, an attention-guided diffusion-contrastive framework that jointly exploits feature-space perturbation, contrastive alignment, and relation-aware spectral attention.}
    \item \rev{We clarify and instantiate the diffusion component as feature-space denoising augmentation rather than full graph generation. The module constructs controlled noisy node-feature views and combines them with contrastive learning to stabilize anomaly-sensitive representations under scarce and imbalanced supervision.}
    \item \rev{We design a relation-aware multi-hop spectral attention module that adaptively integrates hop-level and relation-level cues, reducing the risk of spatial over-smoothing, spectral low-pass bias, and heterogeneous semantic collapse.}
    \item \rev{We expand the empirical evaluation with protocol-consistent recent baselines, training-ratio comparisons, oversampling alternatives, granular module ablations, diffusion-schedule analysis, split stability analysis, and runtime/memory comparisons, providing a more conservative and reproducible assessment of the proposed framework.}
\end{itemize}

\section{Related Work}
% Recent progress in graph-based anomaly detection has yielded Spatial-Based GNN and Spectral-Based GNN approaches, both of which exploit graph structures and features through distinct mechanisms to address varied fraud detection challenges.

\subsection{Spatial-based GNNs}
Spatial GNNs, such as GCN, GraphSAGE, and GIN~\cite{hamilton2017inductive, xu2018powerful,kipf2016semi}, aggregate neighborhood information to capture local structural patterns. In fraud detection, methods including CARE-GNN~\cite{dou2020enhancing}, FRAUDRE~\cite{zhang2021fraudre}, and PC-GNN~\cite{liu2021pick} improve robustness via reinforcement-based thresholding, relation-specific edges, or label-balanced sampling. Subsequent works like GTAN~\cite{xiang2023semi}, Dig-In-GNN~\cite{zhang2024dig}, and ConsisGAD~\cite{chen2024consistency} introduce temporal modeling and contrastive regularization to enhance message quality under label scarcity. 
However, these spatial models rely heavily on the homophily assumption that neighboring nodes share similar semantics, which does not hold in fraud networks where fraudulent nodes are intentionally camouflaged within benign clusters. As a result, minority anomaly signals are easily diluted through message passing, and gradients are dominated by the majority class~\cite{cheng2020graph, tian2023sad}.

\subsection{Spectral-based GNNs}
Spectral approaches analyze graph signals in the frequency domain. GWNN~\cite{xu2019graph} and its variants emphasize low-frequency components to enforce smoothness, while BWGNN~\cite{tang2022rethinking} and AMNet~\cite{chai2022can} attempt to reintroduce mid- and high-frequency information using adaptive kernels and attention mechanisms. Yet, their filtering processes are often biased toward low-pass smoothing, suppressing high-frequency irregularities—the very cues most indicative of fraudulent behavior. When labeled data are scarce, gradients corresponding to minority classes cannot sustain these high-frequency channels, leading to spectral collapse and loss of anomaly relevance~\cite{wu2023splitgnn, guo2024graph}.

\subsection{Heterogeneous GNNs}
Real-world fraud networks involve multiple entity and relation types, such as users, devices, merchants, and transactions~\cite{xu2024revisiting, tang2025fraud, liu2024global}. Models such as HeteroGNN and GHRN~\cite{gao2023addressing, zhang2024heterogeneous} aim to exploit relation-specific semantics via edge-level modeling or structure refinement. However, many approaches still collapse heterogeneous relations into a unified adjacency or homogeneous Laplacian, causing semantic loss. Important relation-driven patterns—like repeated account–device interactions or merchant collusions—are blurred into background noise once aggregated indiscriminately~\cite{liu2018heterogeneous, li2023multi, duan2024dga, hu2024f2gnn}.

\begin{redblock}
Recent studies further extend graph fraud and anomaly detection under sparse supervision and distribution shift. CGAD introduces a contrastive learning-based framework for anomaly detection in attributed networks, improving anomaly-aware representation learning by considering negative-sampling bias and neighborhood information~\cite{wan2024cgad}. ARC explores a generalist graph anomaly detector with in-context learning, aiming to transfer anomaly patterns across graphs using few-shot normal samples~\cite{liu2024arc}. UniGAD unifies node-, edge-, and graph-level anomaly detection through a Rayleigh-quotient-based subgraph sampler and a shared GraphStitch network~\cite{lin2024unigad}. CGNN focuses on graph-based fraud detection with extremely limited labels by combining context-aware category semantics, denoising attention, feature augmentation, and consistency regularization~\cite{li2025context}. These methods are included in the experimental comparison because they are relevant to scarce-label graph anomaly/fraud detection and can be evaluated under our unified 1\% AUC/macro-F1 protocol.

Beyond graph fraud detection, several adjacent representation-learning paradigms are also relevant to ADC-GNN's design scope.
Knowledge condensation distillation studies how to retain compact but informative supervision under limited or redundant knowledge transfer~\cite{li2022knowledge}.
Cross-network embedding methods study transfer and alignment across graph domains where directly shared labels or anchors are sparse~\cite{shen2019network}.
AFSC uses adaptive Fourier-space compression for anomaly-sensitive representation modeling, which is conceptually related to preserving abnormal spectral components~\cite{xu2022afsc}.
InfoBridge-style information-bridge estimation and multimodal bridge modeling provide a broader view of dependency-preserving representation learning under weak supervision~\cite{kholkin2025infobridge}.
The heterogeneous graph learning study by Li et al. further illustrates how cross-modal graph construction and multi-modal completion can support robust representation learning across diverse relation types~\cite{li2024gtp4o}.
These studies are not direct fraud-detection baselines because they target different modalities, tasks, or evaluation protocols; however, they help position ADC-GNN as a task-specific instantiation of robust low-label heterogeneous representation learning.
\end{redblock}

These observations reveal three intertwined challenges in real-world fraud graphs: (i) minority-signal dilution in spatial aggregation, (ii) suppression of fraud-relevant high-frequency components in spectral filtering, and (iii) semantic collapse under heterogeneous relations. Previous studies have addressed each partially—via consistency regularization, edge pruning, or dual-frequency filters—but none jointly overcomes all three limitations under few-shot and imbalanced supervision.

\begin{redblock}
To bridge this gap, we propose ADC-GNN, a unified graph--spectrum framework that integrates feature-space diffusion-guided augmentation, spectral attention, and relation-aware fusion.
Rather than generating new nodes, synthetic transactions, or graph topology, the diffusion component constructs noise-perturbed feature views and regularizes them through denoising and contrastive alignment.
A multi-hop spectral attention mechanism adaptively emphasizes anomaly-related frequency bands while maintaining stability under few-shot settings.
Finally, a relation-aware attention fusion module preserves heterogeneous semantics and reduces information collapse across relation types.
Together, these components form a robust joint embedding space that captures both global and local fraud cues across relational contexts while keeping the scope of the diffusion component technically precise.
\end{redblock}

\section{Problem Definition}
In this section, we formally define the concept of a multi-relational graph and introduce the graph-based fraud detection problem. \\
%\textbf{Graph Modeling.} We define a graph as $G = (V, E, X, Y)$, where $V = \{v_1, v_2, \ldots, v_N\}$ is the set of nodes (with $N = |V|$), $E = \{E_1, E_2, \ldots, E_R\}$ is the set of edges partitioned into $R$ distinct relation types, $X \in \mathbb{R}^{N \times d}$ is the node feature matrix (each node $v_i$ has a $d$-dimensional feature vector), and $Y \in \{0,1\}^N$ is the label vector assigning a binary label to each node. We denote by \(y_i \in \{0,1\}\) the label of node \(v_i\), where \(y_i = 1\) indicates a fraudulent node and \(y_i = 0\) indicates a legitimate (benign) node.\\
\textbf{Graph Modeling.} Let $G = (V, E, X, Y)$ be a graph, where 
$V = \{v_1, v_2, \dots, v_N\}$ is the node set (with $N = |V|$), 
$E = \{E_1, E_2, \dots, E_R\}$ is the edge set partitioned into $R$ relation types, 
$X \in \mathbb{R}^{N \times d}$ is the node feature matrix (each node $v_i$ has a $d$-dimensional feature vector $x_i$), 
and $Y \in \{0,1\}^N$ is the label vector, where $y_i = 1$ indicates a fraudulent node and $y_i = 0$ indicates a legitimate (benign) node, capturing structural and feature information.\\
\textbf{Relational adjacency.}
For each relation type $r\in\{1,\dots,R\}$, let $A^{(r)}\in\{0,1\}^{N\times N}$ be the adjacency of the $r$-th edge subset $E_r$,
and $\tilde A^{(r)}=D^{(r)-\frac12}\!\big(A^{(r)}+I\big)D^{(r)-\frac12}$ its symmetrically normalized version with self-loops.
When the underlying graph is mono-relational, we have $R{=}1$ and the formulation reduces to the standard single-adjacency case.\\
%
%\textbf{Graph-Based Fraud Detection.} Given a graph $G$ as defined above, the goal of graph-based fraud detection is to determine for each node $v_i \in V$ whether it is fraudulent or benign. In other words, this task is formulated as a binary node classification problem, where the objective is to predict the ground-truth label $y_i$ for each node.\\
\textbf{Graph-Based Fraud Detection.} Given graph $G$ as defined above, graph-based fraud detection is formulated as a binary node classification task. The objective is to predict for each node $v_i \in V$ whether it is fraudulent ($y_i=1$) or benign ($y_i=0$), based on relational context.\\
\textbf{Spectral-based Fraud Detection.} These approaches leverage the frequency characteristics of graphs by analyzing signals defined over nodes. A common assumption is that legitimate nodes exhibit smooth behaviors aligned with the graph’s low-frequency spectrum, whereas fraudulent nodes manifest as high-frequency anomalies that deviate from local structural consistency, motivating the use of spectral analysis to expose subtle structural irregularities that are difficult to capture through local aggregation alone.

\begin{figure*}[!t]
    \centering
    \hspace*{-1.5em}
    \includegraphics[width=1.08\linewidth]{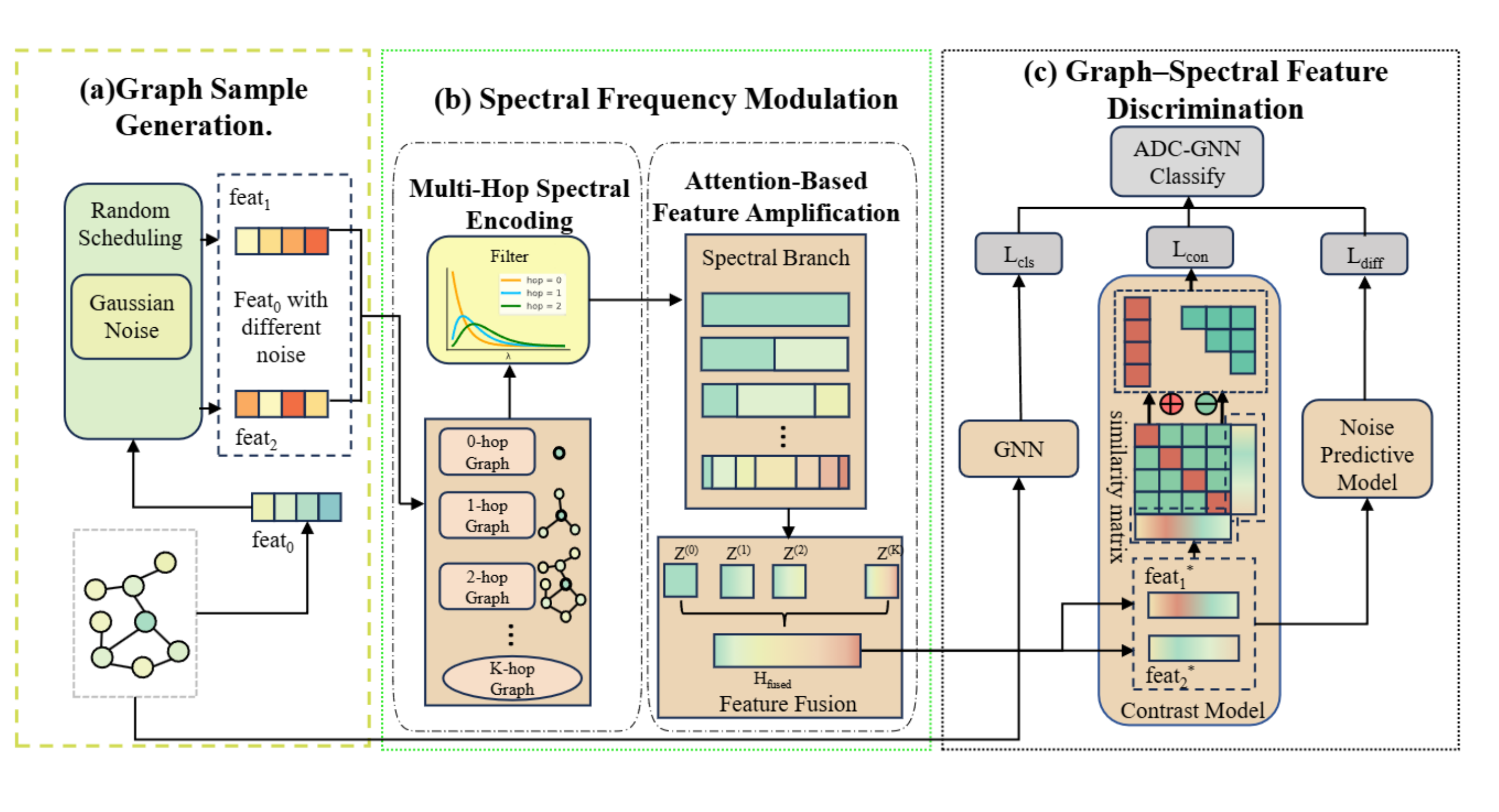}
    \caption{Overall architecture of ADC-GNN.}
    \label{fig:workflow}
\end{figure*}

\section{Methodology}
\begin{redblock}
Our proposed ADC-GNN (Figure~\ref{fig:workflow}) mitigates label scarcity and class imbalance in graph-based fraud detection by integrating feature-space diffusion-guided augmentation with a contrastive, relation-aware spectral GNN backbone~\cite{kipf2016semi}.
As depicted in Figure~\ref{fig:workflow}, the model consists of three linked modules.
(a) Graph sample generation constructs noise-perturbed node-feature views in the feature space under a cosine schedule.
Importantly, this module perturbs node attributes rather than creating new graph topology; thus, it is best interpreted as denoising-based feature augmentation, not as a full graph generative diffusion model.
(b) Spectral frequency modulation converts each noisy view into multi-hop relation-aware spectral representations and uses adaptive attention to emphasize fraud-relevant frequency bands and relations.
(c) Graph--spectral feature discrimination fuses graph- and spectral-domain embeddings and optimizes classification, denoising, and contrastive objectives to improve anomaly separability under scarce supervision.
\end{redblock}

\subsection{Graph Sample Generation}

\begin{redblock}
In practical applications, the high cost of annotation results in scarce labeled samples, limiting supervised learning and weakening the model's ability to recognize rare anomalous patterns.
Inspired by diffusion models in computer vision~\cite{nichol2021improved, dhariwal2021diffusion}, we employ a cosine-scheduled noise injection mechanism to perturb node features and construct multiple noisy views.
However, unlike full graph diffusion models that learn a reverse process for generating both node attributes and graph topology, our module does not add or remove edges.
This design choice is deliberate: in transaction graphs, edges usually correspond to observed business events or verified relations, and arbitrary topology perturbation may introduce invalid transactions or destroy relation semantics.
Therefore, our diffusion module is used as a feature-space denoising augmentation strategy.
It exposes the encoder to controlled perturbations of node attributes and relies on contrastive learning to preserve node identity while improving representation robustness under scarce supervision.
\end{redblock}

Specifically, we define the cumulative retention factor at diffusion step $t$, where $T$ denotes the total number of diffusion steps~\cite{nichol2021improved}, and $s$ is a small positive constant introduced to prevent the process from starting at zero—an issue that could otherwise cause instability by forcing the model to predict noise when none exists. At each step $t$, the node feature $x_0$ is perturbed as:
\begin{equation}
\begin{aligned}
  \bar{\alpha}_t 
    &= \prod_{i=0}^t \alpha_i 
    = \cos^2\!\Bigl(\frac{\tfrac{t}{T} + s}{1 + s}\,\frac{\pi}{2}\Bigr), \\
  x_t
    &= \sqrt{\bar{\alpha}_t}\,x_0 + \sqrt{1 - \bar{\alpha}_t}\,\epsilon,
      \quad \epsilon \sim \mathcal{N}(0, I),
\end{aligned}
\end{equation}
\begin{redblock}
Here, $\sqrt{\bar\alpha_t}$ controls the proportion of the original features retained, while $\sqrt{1-\bar\alpha_t}$ governs the magnitude of the injected noise.
As $t$ increases, the retained feature component gradually decreases and the stochastic component increases, producing a spectrum of node-feature views from near-original to strongly perturbed.
The informative signal does not come from Gaussian noise itself, but from the constraint that two independently perturbed views of the same node should remain close in the learned representation while representations of different nodes are separated through the contrastive objective in Section~\ref{sec:graph_spectral_discrimination}.
Thus, the diffusion-guided views act as structured regularization for node representations rather than as random synthetic fraud samples.
In practice, the schedule length $T=1000$ specifies the discretization of the noise schedule.
During training, noisy views can be sampled directly from $q(x_t|x_0)$ at selected timesteps without executing an iterative 1000-step reverse chain, and inference uses the trained encoder/classifier without diffusion sampling, which keeps deployment latency comparable to standard GNN inference.
\end{redblock}

\subsection{Spectral Frequency Modulation}
\textbf{Multi-Hop Spectral Encoding.}
To transform graph data into the spectral domain, we first preprocess the features through a three-stage pipeline. We apply two linear transformations, each followed by a non-linear activation $\phi$ (ReLU)~\cite{glorot2011deep}, and then apply dropout~\cite{srivastava2014dropout}. This preprocessing yields a smoother, normalized, and expressive feature representation $h_{\mathrm{pre}}$, which serves as the input for subsequent spectral encoding:
\begin{equation}
\begin{aligned}
h^{(1)} &= \phi\bigl(W^{(1)} x + b^{(1)}\bigr),\\
h^{(2)} &= \phi\bigl(W^{(2)}\,h^{(1)} + b^{(2)}\bigr),\\
\tilde{h}^{(2)} &= \mathrm{dropout}\bigl(h^{(2)}\bigr).
\end{aligned}
\end{equation}

To construct robust, scalable spectral representations, we employ a unified strategy known as Multi-Hop Spectral Encoding. This approach leverages an $n$-order polynomial expansion to build $n$ parallel branches, where each branch captures graph signals at a specific hop scale and encodes local structural information~\cite{xu2018powerful,defferrard2016convolutional}. Lower-order terms emphasize immediate neighborhood structures, while higher-order terms progressively incorporate long-range dependencies. By aggregating these multiple branches, the model is able to balance local smoothness with global anomaly cues, thereby avoiding over-smoothing and preserving discriminative high-frequency information. The resulting multi-scale features are then fused into a compact multi-band spectral representation, which retains both fine-grained local semantics and coarse-grained global patterns essential for fraud detection.

\noindent\textit{Relation-aware polynomial filtering.}
Within the multi-hop spectral module, each branch $k$ applies relation-specific polynomial filters of order $d$ with learnable coefficients $\alpha^{(k,r)}_i$ to aggregate information from $0$-hop to $d$-hop neighborhoods of relation $r$:
\begin{equation}
z^{(k,r)} = \sum_{i=0}^{d} \alpha^{(k,r)}_{i}\, \bigl(\tilde{A}^{(r)}\bigr)^{i}\, h_{\mathrm{pre}}, 
\qquad r=1,\dots,R,
\end{equation}
where $\tilde{A}^{(r)}$ denotes the normalized adjacency matrix of relation $r$. 

\noindent\textit{Relation attention fusion.}
To prevent collapsing heterogeneous edge semantics, we learn attention weights $\beta^{(k)}_{i,r}$ for each node $i$ over relations:
\begin{equation}
\beta^{(k)}_{i,r} = \frac{\exp\!\Big(q_k^\top \tanh\!\big(W_k [\,z^{(k,r)}_{i}; h_{\mathrm{pre},i}\,]\big)\Big)}
{\sum_{\ell=1}^{R} \exp\!\Big(q_k^\top \tanh\!\big(W_k [\,z^{(k,\ell)}_{i}; h_{\mathrm{pre},i}\,]\big)\Big)},
\end{equation}
where $\sum_{r=1}^R \beta^{(k)}_{i,r} = 1$. The branch output for node $i$ is then obtained as
\begin{equation}
z^{(k)}_{i} = \sum_{r=1}^{R} \beta^{(k)}_{i,r}\, z^{(k,r)}_{i}.
\end{equation}

\begin{redblock}
Since $z^{(k)}$ has already integrated all $R$ relation types through the relation-attention fusion in Eq.~(5), 
the subsequent stacking across branches does not explicitly maintain a separate relation dimension.
\end{redblock} 
Finally, stacking the outputs of all $K$ branches yields a fused tensor:
\begin{equation}
\mathbf{H}_{\mathrm{fused}} = \mathrm{stack}\bigl(z^{(1)}, z^{(2)}, \dots, z^{(K)}\bigr)
\;\in\;\mathbb{R}^{N\times K\times F},
\end{equation}
where $K$ is the number of branches and $F$ is the feature dimensionality of each branch’s output. 
The resulting fused tensor $\mathbf{H}_{\mathrm{fused}}$ thus provides a unified representation of multi-scale structural information while preserving relational heterogeneity through attention-based fusion.

%%%%%%%%%%%%%%%%%%%%%%%%%%%%%%%%%%%%%

%%%%%%%%%%%%%%%%%%%%%%%%%%%%%%%%%%%%%

\textbf{Attention-Based Feature Amplification.}
%To highlight salient features, we propose a branch attention mechanism that adaptively assigns attention weights to each node’s multi-hop feature channels. We first stack the feature vectors from \(K\) selected branches of \(N\) nodes into a three-dimensional tensor and then flatten it into a two-dimensional matrix \(\mathbf{H}_{\mathrm{embed}}\), where row \((i,j)\) corresponds to branch \(j\) of node \(i\). Next, we apply a shared fully-connected layer to each row to compute the unnormalized attention score \(s_{i,j}\).
%
%~\cite{velivckovic2017graph}:
%\begin{equation}
%\begin{aligned}
%  &\mathbf{H}_{\mathrm{embed}}[i \cdot K + j, :] = \mathbf{H}_{\mathrm{fused}}[i, j, :],\\
%  &s_{i,j} = \mathrm{FC}\bigl(\mathbf{H}_{\mathrm{embed}}[i \cdot K + j, :]\bigr).
%\end{aligned}
%\end{equation}
%By reshaping these scores back into matrix form, we obtain
%\begin{equation}
%\begin{aligned}
%  S \in \mathbb{R}^{N\times K},
%  S_{i,:} = [s_{i,1}, s_{i,2}, \dots, s_{i,K}],
%\end{aligned}
%\end{equation}
%contains the scores of the $K$ branches for node $i$. We then normalize these scores into attention weights by applying a softmax function across the $K$ branches of each node~\cite{hu2018squeeze}:
%\begin{equation}
%\alpha_{i,j} = \frac{\exp(s_{i,j})}{\sum_{l=1}^{K}\exp(s_{i,l})}.
%\end{equation}
To highlight salient features, we propose a branch attention mechanism that adaptively assigns attention weights to each node’s multi-hop feature channels. Different hop scales capture distinct structural semantics, and directly averaging them may dilute informative signals. To address this, we reweight branch features based on their relative importance.

Formally, we stack the feature vectors from $K$ branches of $N$ nodes into a three-dimensional tensor $\mathbf{H}_{\mathrm{fused}}$, and flatten it into a two-dimensional matrix $\mathbf{H}_{\mathrm{embed}}$, where row $(i,j)$ corresponds to branch $j$ of node $i$. A fully-connected layer is applied to compute the unnormalized attention score $s_{i,j}$~\cite{velivckovic2017graph}:
\begin{equation}
\begin{aligned}
  &\mathbf{H}_{\mathrm{embed}}[i \cdot K + j, :] = \mathbf{H}_{\mathrm{fused}}[i, j, :],\\
  &s_{i,j} = \mathrm{FC}\bigl(\mathbf{H}_{\mathrm{embed}}[i \cdot K + j, :]\bigr).
\end{aligned}
\end{equation}

Reshaping these scores back into matrix form yields
\begin{equation}
S \in \mathbb{R}^{N\times K}, \quad
S_{i,:} = [s_{i,1}, s_{i,2}, \dots, s_{i,K}],
\end{equation}
which contains the branch-level scores for node $i$. We then normalize them with a softmax across branches~\cite{hu2018squeeze}:
\begin{equation}
\alpha_{i,j} = \frac{\exp(s_{i,j})}{\sum_{l=1}^{K}\exp(s_{i,l})}.
\end{equation}
Here, $\alpha_{i,j}$ denotes the normalized attention weight, guiding the model to emphasize more discriminative hop-level features for each node.

%The weights $\alpha_{i,j}$ reflect the importance of each hop-scale feature for node $i$, guiding the model to focus on more discriminative branches.

%When computing the final node representation \(\hat h_i\), we introduce an explicit mean-fusion weight alongside the attention weight, and integrate both with a residual projection of the preprocessed feature. Concretely, let \(z_i^{(j)}\) be the output of branch \(j\) for node \(i\). We define
%These weights $\alpha_{i,j}$ reflect the importance of the $j$-hop features for node $i$, guiding the model to focus on more discriminative spectral branches. In computing the final node representation $\hat{h}_i$, we also integrate a mean-fusion component and a residual connection. Let $z_i^{(j)}$ denote the output of branch $j$ for node $i$. We define:
These weights $\alpha_{i,j}$ indicate the relative importance of the $j$-hop features for node $i$, encouraging the model to attend to more informative spectral branches. To obtain the final node representation $\hat{h}_i$, we integrate the attention-weighted branch outputs with a mean-fusion term for stability, as well as a residual projection to retain the original feature information. Let $z_i^{(j)}$ denote the output of branch $j$ for node $i$. We define:
%These weights $\alpha_{i,j}$ represent the relative importance of the $j$-hop features for node $i$, guiding the model to selectively emphasize more discriminative spectral branches while down-weighting less informative ones. To construct the final node representation $\hat{h}_i$, we combine the attention-weighted aggregation of branch outputs with a mean-fusion component that stabilizes training, together with a residual projection that preserves the original preprocessed features. Let $z_i^{(j)}$ denote the output of branch $j$ for node $i$. We define:

%These weights $\alpha_{i,j}$ quantify the relative importance of the $j$-hop features for node $i$, guiding the model to emphasize more discriminative spectral branches. To form the final node representation $\hat{h}_i$, we combine three complementary components: (i) an attention-weighted aggregation over hop-specific outputs, (ii) a mean-fusion term that provides a balanced baseline across branches, and (iii) a residual projection of the preprocessed feature to preserve original structural information. Concretely, let $z_i^{(j)}$ denote the output of branch $j$ for node $i$. We define:

\begin{equation}
A_i = \sum_{j=1}^K \alpha_{i,j}\,z_i^{(j)},\quad
M_i = \frac{1}{K}\sum_{j=1}^K z_i^{(j)},\quad
R_i = W_{\mathrm{res}}\,h_{\mathrm{pre},i},
\end{equation}
where \(A_i\) is the attention-weighted fusion, \(M_i\) is the mean of branch outputs, and \(R_i\) is the residual projection of the preprocessed feature \(h_{\mathrm{pre}}\). We then compute an adaptive fusion coefficient \(\beta_i\) via a small multilayer perceptron:
\begin{equation}
\beta_i = \sigma\bigl(\mathrm{MLP}([A_i; M_i])\bigr),\quad
\mu_i = 1 - \beta_i,
\end{equation}
where \(\sigma\) denotes the sigmoid function and \(\mu_i\) is the corresponding mean-fusion weight. Finally, the three components are combined as
\begin{equation}
\hat f_i = \beta_i\,A_i + \mu_i\,M_i + \gamma\,R_i,
\end{equation}
with \(\gamma\) a hyperparameter controlling the contribution of the residual term. This formulation allows the model to balance attention, mean pooling, and original structural information in a fully adaptive manner.~\cite{hu2018squeeze,bhatia2020midas}

\begin{algorithm}[!t]
\caption{\rev{Training Process of ADC-GNN with Timestep Sampling}}
\label{alg:simplified}
\begin{algorithmic}[1]
\Require Graph $G=(V,E,X,Y)$; diffusion schedule length $T$; branches $K$
\Ensure Fused features $\hat{\mathbf{f}}$; total loss $\mathcal{L}_{\mathrm{total}}$

\Statex \rev{\textbf{// Diffusion-guided feature-view construction}}
\State \rev{Sample a timestep $t \sim \mathrm{Uniform}(\{1,\ldots,T\})$ for the current mini-batch}
\State \rev{Sample two independent noises $\epsilon_1,\epsilon_2 \sim \mathcal{N}(0,I)$}
\State \rev{Construct two noisy feature views $x_t^{(1)}$ and $x_t^{(2)}$ directly from $q(x_t|x_0)$}

\Statex \rev{\textbf{// Relation-aware spectral frequency modulation}}
\State Obtain $h_{\mathrm{pre}}$ from $x_t^{(1)}$ and $x_t^{(2)}$
\For{$k \gets 1$ \textbf{to} $K$}
    \State Compute relation-aware polynomial features $z^{(k)}$
\EndFor

\Statex \rev{\textbf{// Feature fusion and optimization}}
\For{each node $i$ in the current training batch $\mathcal{B}$}
    \State Obtain $\bar{\mathbf{f}}_i$ via attention, mean, and residual fusion of $\{z_i^{(k)}\}_{k=1}^K$
\EndFor
\State Compute $\mathcal{L}_{\mathrm{cls}}$ on labeled nodes $\mathcal{V}_L \cap \mathcal{B}$, and compute $\mathcal{L}_{\mathrm{diff}}$ and $\mathcal{L}_{\mathrm{con}}$ on the sampled feature views
\State Obtain total loss $\mathcal{L}_{\mathrm{total}} \gets \lambda_{\mathrm{cls}}\mathcal{L}_{\mathrm{cls}} + \lambda_{\mathrm{diff}}\mathcal{L}_{\mathrm{diff}} + \lambda_{\mathrm{con}}\mathcal{L}_{\mathrm{con}}$
\end{algorithmic}
\end{algorithm}

\subsection{Graph--Spectral Feature Discrimination}\label{sec:graph_spectral_discrimination}

\rev{To effectively leverage local neighborhood information, we feed the original input features into the GNN backbone to obtain the logits $\hat{y}_i$.} These logits are supervised using the ground-truth labels $y_i \in \{0, 1\}$ through the binary cross-entropy loss, which yields the classification loss $\mathcal{L}_{\mathrm{cls}}$.
%We feed the original input features into the GNN backbone to obtain  logits $\hat{y}i$ for each node. These predictions are supervised with the binary cross-entropy loss $L{cls}$:
%
\begin{redblock}
Let $\mathcal{V}_L$ denote the labeled training nodes.
To avoid using validation and test labels during optimization, the supervised classification loss is computed only over $\mathcal{V}_L$:
\begin{equation}
\mathcal{L}_{\mathrm{cls}} =
-\frac{1}{|\mathcal{V}_L|}
\sum_{i\in\mathcal{V}_L}
\left[
y_i \log(\hat{y}_i) + (1-y_i)\log(1-\hat{y}_i)
\right].
\end{equation}
Here, $y_i$ is the true label of labeled training node $v_i$, and $\hat{y}_i$ is the  fraud probability output by the model.
Unlabeled, validation, and test nodes may be used through their graph structure and node features in the transductive graph setting, but their labels are never used in $\mathcal{L}_{\mathrm{cls}}$.
\end{redblock}

To enhance the discriminative power of the learned features, we independently sample two noise realizations, $\epsilon_1$ and $\epsilon_2$, for each node’s original feature vector $x_0$, thereby constructing two noisy views at diffusion step $t$:
\begin{equation}
x_t^{(1)} = q(x_0, \epsilon_1), 
\quad
x_t^{(2)} = q(x_0, \epsilon_2).
\end{equation}

These two noisy features are then processed by the Spectral Frequency Modulation module for feature enhancement. The resulting fused representations $\mathbf{f}^{(1)}$ and $\mathbf{f}^{(2)}$ are subsequently input into the Noise Predictive Model. For each view, the model predicts the corresponding noise, and the diffusion loss is computed as:
\begin{equation}
\begin{aligned}
(\hat{\epsilon}^{(i)},\, \mathbf{f}^{(i)}) 
&= \mathcal{M}(\mathbf{f}^{(i)},\, t), \quad i = 1, 2, \\
\mathcal{L}_{\mathrm{diff}} 
&= \frac{1}{2}\sum_{i=1}^{2}\mathrm{MSE}(\hat{\epsilon}^{(i)},\, \epsilon_i).
\end{aligned}
\end{equation}
\rev{where $\mathcal{M}(\cdot,t)$ denotes the lightweight noise-prediction head and $\mathrm{MSE}(\cdot,\cdot)$ denotes the mean squared error.
The denoising loss encourages the model to retain perturbation-aware information, but by itself it does not enforce class separation; this is why we separately introduce the contrastive objective below.}

Note that this loss supervises only the denoising capability and does not involve contrastive supervision. To further encourage feature separability, we introduce a contrastive loss $\mathcal{L}_{\mathrm{con}}$, which pulls together the embeddings of the same node under different noise perturbations and pushes apart those of different nodes.
\begin{redblock}
In implementation, the contrastive loss is optimized on a mini-batch $\mathcal{B}$ with sampled negatives rather than by materializing all $N^2$ node pairs:
\begin{equation}
\mathcal{L}_{\mathrm{con}}
= -\,\frac{1}{|\mathcal{B}|}
\sum_{i\in\mathcal{B}}
\log
\frac{
  \exp\!\bigl(\mathrm{sim}(\bar f_i^{(1)},\,\bar f_i^{(2)})/\tau\bigr)
}{
  \sum_{j\in\mathcal{B}}
  \exp\!\bigl(\mathrm{sim}(\bar f_i^{(1)},\,\bar f_j^{(2)})/\tau\bigr)
}\,.
\end{equation}
\end{redblock}
where $\mathrm{sim}(\cdot,\cdot)$ denotes cosine similarity (implemented as the dot product between $\ell_2$-normalized vectors), and $\tau$ is a temperature parameter controlling distribution sharpness.

To further emphasize anomalous samples, we define a positive pair for each sample $i$ as $(\bar{\mathbf{f}}_i^{(1)},\, \bar{\mathbf{f}}_i^{(2)})$, where both embeddings are generated from independent noise perturbations of the same input. In contrast, for $j \ne i$, the pair $(\bar{\mathbf{f}}_i^{(1)},\, \bar{\mathbf{f}}_j^{(2)})$ is treated as a negative pair.

Following~\cite{chen2020simple}, we compute a cross-entropy loss for each sample $i$ that maximizes the similarity of its positive pair while minimizing the similarities of its negative pairs. Averaging over all samples yields the contrastive loss $\mathcal{L}_{\mathrm{con}}$. This loss encourages the model to align the feature representations of the same node under different noise views, while pushing apart those of different nodes. Such contrastive regularization helps the model better capture fine-grained variations of anomalous samples throughout the denoising process. We define the total training objective as the weighted sum of the contrastive loss, diffusion loss, and classification loss:
\begin{equation}
\mathcal{L}_{\mathrm{total}} 
= \lambda_{\mathrm{con}} \,\mathcal{L}_{\mathrm{con}} 
+ \lambda_{\mathrm{diff}} \,\mathcal{L}_{\mathrm{diff}} 
+ \lambda_{\mathrm{cls}} \,\mathcal{L}_{\mathrm{cls}}.
\end{equation}
Here, $\lambda_{\mathrm{con}}, \lambda_{\mathrm{diff}}, \lambda_{\mathrm{cls}} \geq 0$ are trade-off coefficients that balance the contributions of the three objectives. 
Specifically, $\lambda_{\mathrm{con}}$ controls the strength of feature alignment under contrastive learning, 
$\lambda_{\mathrm{diff}}$ determines the relative importance of denoising guidance in the diffusion process, 
and $\lambda_{\mathrm{cls}}$ adjusts the emphasis placed on supervised node classification.
This composite objective integrates three complementary learning signals: 
(i) feature alignment via $\mathcal{L}_{\mathrm{con}}$, 
(ii) denoising guidance via $\mathcal{L}_{\mathrm{diff}}$, and 
(iii) supervised classification via $\mathcal{L}_{\mathrm{cls}}$. 
Together, these weighted components drive convergence and ensure that subtle but informative anomalous patterns are preserved throughout training.

\section{Experiments}
\label{sec:experiments}
In this section, we evaluate ADC-GNN on three public fraud-detection benchmarks and report the proprietary CM dataset as a private deployment-style case study. The public datasets provide the primary reproducible evidence, while CM is used to illustrate applicability in a real telecom risk-control scenario. To keep the evaluation self-contained, we move the core experimental clarifications and additional experimental analyses into this section, including dataset semantics, CM dataset disclosure, split protocol, implementation fairness, stability analysis, training-ratio comparison, oversampling comparison, diffusion-schedule analysis, complexity discussion, and granular ablation interpretation. The appendix is retained only for supplementary dataset and reproducibility notes. We answer the following research questions:

\textbf{RQ1}: How does ADC-GNN compare with graph fraud detection methods under conditions of extremely limited labels and severe class imbalance?

\textbf{RQ2}: What is the individual contribution of each key component of ADC-GNN---specifically the contrastive diffusion module and the attention mechanism---to the model's aggregate performance?

\textbf{RQ3}: How sensitive is ADC-GNN's detection performance to the choice of critical training hyperparameters and diffusion schedules?

\textbf{RQ4}: How do the submodules within ADC-GNN coordinate to progressively reconstruct and enhance discriminative graph representations under few-shot and imbalanced settings?

\rev{\textbf{RQ5}: Is ADC-GNN robust under different random few-shot splits, different training ratios, and classical oversampling alternatives?}

\rev{\textbf{RQ6}: What computational overhead does ADC-GNN introduce compared with representative graph fraud detection baselines?}

\subsection{Experimental Settings}

\textbf{Datasets.} We conduct the main reproducible experiments on three public datasets: Amazon~\cite{rayana2015collective}, YelpChi~\cite{mcauley2013amateurs}, and T-Finance~\cite{tang2022rethinking}. We additionally include CM as a proprietary telecom transaction fraud dataset. Since CM cannot be independently released, it is reported as a private case study with schema-level disclosure rather than as the main reproducibility anchor. Table~\ref{tab:dataset_stats_selected} reports the dataset statistics.

Amazon is a public review-spam benchmark in which nodes correspond to reviewer or user entities and edges are derived from review-related relational links released by the benchmark. These relations encode reviewer-item or review-behavior associations that are widely used in graph-based fraud detection. The node features are behavioral and metadata-derived descriptors, and the labels indicate whether the corresponding reviewer or review-related entity is fraudulent or benign. YelpChi is another public review-spam benchmark, where nodes correspond to reviewers and edges are constructed from review-, product-, and behavior-based co-occurrence relations. Its features describe user-review behavioral patterns, and its labels are public spam-review labels. T-Finance is a public financial fraud detection benchmark. Its nodes correspond to account or transaction-related financial entities, and its edges describe benchmark-defined financial interaction relations. The released anonymized financial behavioral features and public fraud labels are used under the same few-shot split protocol as the other datasets.

\begin{redblock}
The CM dataset is collected from a company-owned telecom risk-control scenario over approximately 11 months. It contains 61{,}582 nodes, 39{,}381{,}037 edges, 14 de-identified node features, and a fraud ratio of 3.76\%. In CM, each node denotes an anonymized telecom account or transaction-related entity. Raw account identifiers, phone numbers, device identifiers, transaction identifiers, transaction contents, and business-rule thresholds cannot be released due to privacy, compliance, and contractual restrictions. Edges are constructed from verified interaction evidence, including shared-device links, shared-contact or identity-attribute links, transaction co-occurrence links, and regional-temporal co-occurrence links. The node features are de-identified numerical or categorical risk descriptors summarizing transaction frequency, account behavior, regional information, device or identity risk, and historical interaction statistics. Positive labels are obtained from the internal risk-control system and manual verification records, while negative labels correspond to entities not verified as fraudulent during the observation and verification window. To reduce temporal leakage, all features are generated only from information available before the label-verification time. Therefore, the CM experiment controls feature-time leakage, although it should not be interpreted as a fully prospective temporal deployment evaluation.

A concrete CM example is a telecom risk-control case in which several newly registered accounts repeatedly access services through the same device fingerprint, share similar contact attributes, conduct transactions within a short regional-temporal window, and interact with overlapping counterparties. Each account may appear normal when considered independently, and each single behavior may be insufficient to confirm fraud. However, their graph-level co-occurrence can form a suspicious local pattern: the accounts may share the same device environment, show synchronized regional activity bursts, and repeatedly interact with a small set of common transaction entities. Such cases are difficult for purely feature-based detectors because the anomalous signal is distributed across multiple weak relational cues. ADC-GNN is designed to capture these weak but correlated signals by jointly modeling feature perturbation robustness, relation-aware neighborhood structure, and multi-hop spectral information. This example is provided only to clarify the graph construction logic; exact business rules, thresholds, and private transaction contents are not disclosed.
\end{redblock}

\newcolumntype{L}{>{\raggedright\arraybackslash}X}

\setlength{\tabcolsep}{5pt} % 缩小列间距
\begin{table}[!t]
  \centering
  %\small
  \resizebox{\columnwidth}{!}{
  \begin{tabular}{lcccc}
    \toprule
    \textbf{Dataset} & \textbf{Amazon} & \textbf{YelpChi} & \textbf{T-Finance} & \textbf{CM} \\
    \hline
    \textbf{Nodes}      & 11,944    & 45,954     & 39,357      & 61,582 \\
    \textbf{Edges}      & 4,398,392 & 3,846,979  & 21,222,543  & 39,381,037 \\
    \textbf{Features}   & 25        & 32         & 10          & 14 \\
    \textbf{Fraud (\%)} & 6.87      & 14.53      & 4.58        & 3.76 \\
    \bottomrule
  \end{tabular}
  }
  \caption{Statistics of the selected four datasets.}
  \label{tab:dataset_stats_selected}
\end{table}

\begin{table*}[t]
\centering
\scriptsize
\setlength{\tabcolsep}{3pt}
\renewcommand{\arraystretch}{1.08}
\resizebox{\textwidth}{!}{%
\begin{tabular}{lcccccccc}
\toprule
\textbf{Method}
& \multicolumn{2}{c}{\textbf{Amazon}}
& \multicolumn{2}{c}{\textbf{YelpChi}}
& \multicolumn{2}{c}{\textbf{T-Finance}}
& \multicolumn{2}{c}{\textbf{CM}} \\
\cmidrule(lr){2-3}\cmidrule(lr){4-5}\cmidrule(lr){6-7}\cmidrule(lr){8-9}
& \textbf{AUC} & \textbf{MF1}
& \textbf{AUC} & \textbf{MF1}
& \textbf{AUC} & \textbf{MF1}
& \textbf{AUC} & \textbf{MF1} \\
\midrule
ChebyNet        & 0.8821 & 0.8741 & 0.7422 & 0.6363 & 0.8753 & 0.7918 & 0.8161 & 0.7309 \\
GWNN            & 0.8769 & 0.8899 & 0.7658 & 0.6463 & 0.8734 & 0.7263 & 0.8442 & 0.6654 \\
GIN             & 0.8265 & 0.7281 & 0.6646 & 0.5872 & 0.7004 & 0.6021 & 0.6912 & 0.5412 \\
CARE-GNN        & 0.8869 & 0.5953 & 0.7448 & 0.5874 & 0.9050 & 0.7332 & 0.8258 & 0.6723 \\
PC-GNN          & 0.9148 & 0.8736 & 0.7523 & 0.5498 & 0.9076 & 0.6206 & 0.8384 & 0.5597 \\
BW-GNN          & 0.9075 & 0.9109 & 0.7467 & \underline{0.6726} & 0.9275 & \underline{0.8573} & 0.8583 & 0.7374 \\
COFD            & 0.9012 & 0.8795 & 0.7026 & 0.5907 & 0.9010 & 0.8322 & 0.8733 & 0.7357 \\
GTAN            & 0.9079 & 0.8754 & 0.7571 & 0.6017 & 0.9074 & 0.8028 & 0.8782 & 0.7696 \\
GHRN            & 0.9276 & \underline{0.9129} & 0.7833 & 0.6608 & 0.9304 & 0.8305 & 0.8424 & 0.7450 \\
DIG-In-GNN      & \underline{0.9279} & 0.8447 & 0.7368 & 0.6386 & 0.9131 & 0.6851 & \underline{0.8828} & 0.5949 \\
ConsisGAD       & 0.9201 & 0.9054 & 0.8000 & 0.6321 & \underline{0.9316} & 0.8513 & 0.8815 & 0.7720 \\
\rev{CGAD}      & \rev{0.9186} & \rev{0.9049} & \rev{0.7936} & \rev{0.6584} & \rev{0.9228} & \rev{0.8410} & \rev{0.8796} & \rev{\underline{0.7724}} \\
\rev{ARC}       & \rev{0.9268} & \rev{0.9076} & \rev{\underline{0.8003}} & \rev{0.6688} & \rev{0.9272} & \rev{0.8493} & \rev{0.8788} & \rev{0.7605} \\
\rev{UniGAD}    & \rev{0.9242} & \rev{0.9122} & \rev{0.7968} & \rev{0.6635} & \rev{0.9312} & \rev{0.8569} & \rev{0.8819} & \rev{0.7692} \\
\rev{CGNN}      & \rev{0.9275} & \rev{0.9097} & \rev{0.7986} & \rev{0.6691} & \rev{0.9308} & \rev{0.8549} & \rev{0.8824} & \rev{0.7708} \\
\midrule
\textbf{ADC-GNN} & \textbf{0.9501} & \textbf{0.9230} & \textbf{0.8316} & \textbf{0.6949} & \textbf{0.9503} & \textbf{0.8924} & \textbf{0.9397} & \textbf{0.8269} \\
\bottomrule
\end{tabular}%
}
\caption{\rev{Experimental results at a 1\% training ratio under the same fixed split and random seed (seed=72) as the original submission. Original baseline results are preserved, and protocol-consistent recent graph anomaly/fraud baselines are additionally included. Bold values indicate the best overall results, and underlined values indicate the strongest non-ADC-GNN baseline in each column. Amazon, YelpChi, and T-Finance are the main public benchmarks. CM is reported as a proprietary private case study and is not used as the primary reproducibility anchor.}}
\label{tab:four_datasets_best}
\end{table*}

\begin{redblock}
\noindent\textbf{Split protocol.}
For each dataset, we construct few-shot splits using stratified random sampling according to the node labels, such that the labeled training set approximately preserves the original class distribution.
In the 1\% setting, 1\% of all nodes are used as labeled training nodes, while the remaining nodes are divided into validation and test sets with a validation-to-test ratio of 1:2.
The main comparison follows the same fixed split with random seed 72 as in the original submission to ensure direct comparability with the reported baseline results.
For experiments that require stability analysis, we further repeat the sampling procedure with five independent random seeds: 72, 123, 2024, 3407, and 7777.

Under the 1\% split with seed 72, the training sets contain 119 nodes with 8 fraudulent nodes for Amazon, 459 nodes with 67 fraudulent nodes for YelpChi, 393 nodes with 18 fraudulent nodes for T-Finance, and 615 nodes with 23 fraudulent nodes for CM.
The validation sets contain 3,942, 15,165, 12,988, and 20,322 nodes for Amazon, YelpChi, T-Finance, and CM, respectively.
The corresponding test sets contain 7,883, 30,330, 25,976, and 40,645 nodes, respectively.
\end{redblock}

\begin{table*}[!t]
\centering
\scriptsize
\setlength{\tabcolsep}{6pt}
\renewcommand{\arraystretch}{1.12}
\resizebox{0.72\textwidth}{!}{%
\begin{tabular}{lccc}
\toprule
\textbf{Dataset} & \textbf{Seeds} & \textbf{AUC (\%)} & \textbf{Macro-F1 (\%)} \\
\midrule
\rev{Amazon}    & \rev{5} & \rev{$95.24{\pm}1.25$} & \rev{$92.15{\pm}1.21$} \\
\rev{YelpChi}   & \rev{5} & \rev{$82.89{\pm}0.32$} & \rev{$69.67{\pm}0.26$} \\
\rev{T-Finance} & \rev{5} & \rev{$95.22{\pm}1.36$} & \rev{$88.46{\pm}2.16$} \\
\rev{CM}        & \rev{5} & \rev{$94.21{\pm}1.31$} & \rev{$81.92{\pm}0.82$} \\
\bottomrule
\end{tabular}%
}
\caption{\rev{Five-seed stability analysis of ADC-GNN under the 1\% few-shot protocol. Values are reported in percentage points as mean$\pm$standard deviation over five independent seeds \{72, 123, 2024, 3407, 7777\}. The reported metrics are restricted to AUC and Macro-F1 to remain consistent with the main comparison in Table~\ref{tab:four_datasets_best}. Amazon, YelpChi, and T-Finance are public benchmark datasets, while CM is additionally included as a proprietary real-world case-study dataset. This table is interpreted as a split-stability diagnostic rather than as a replacement for the fixed seed-72 main comparison in Table~\ref{tab:four_datasets_best}.}}
\label{tab:multiseed_stability}
\end{table*}

\textbf{Metrics.} We use two widely adopted metrics: Macro-F1 and AUC. Macro-F1 computes the harmonic mean of precision and recall calculated for each class. AUC measures the model’s ability to rank positive instances higher than negative instances, regardless of any specific decision threshold.

\textbf{Parameter Setting.}
We set the number of diffusion steps to 1000 with a cosine scheduler and an initial offset of $8 \times 10^{-3}$. For contrastive learning, the contrastive loss weight $\lambda$ is 0.1. The hidden dimension is 64, polynomial order is 2, and dropout is 0.2. \rev{The default training set consists of 1\% of all nodes. After removing the labeled training nodes, the remaining nodes are divided into validation and test sets with a validation-to-test ratio of 1:2. In other words, the validation set contains approximately one third of the remaining nodes, while the test set contains approximately two thirds.} The maximum number of training epochs is 100 for Amazon and YelpChi, and 500 for T-Finance and CM due to their more complex relational structures. The learning rate is fixed at 0.01. 
For direct comparability with the original submission, Table~\ref{tab:four_datasets_best} preserves the fixed seed-72 results. To address split stability, Table~\ref{tab:multiseed_stability} reports five-seed reruns under the same 1\% few-shot protocol. Amazon, YelpChi, and T-Finance serve as public reproducibility anchors, while CM is additionally included as a proprietary real-world case-study dataset. Since the raw CM data cannot be released, its reproducibility is supported through schema-level documentation, aggregate statistics, split-generation code, model configurations, and evaluation scripts.
\begin{redblock}
We adopt the cosine schedule as a practical noise schedule commonly used in diffusion-style perturbation processes.
Rather than treating it as a task-specific optimality proof, we use it to obtain a smoothly decaying retention factor that avoids an abrupt transition from clean to heavily corrupted node features.
This is important in few-shot fraud graphs because overly weak perturbations provide limited regularization, whereas overly strong perturbations may erase already scarce minority-class cues.
In our implementation, the schedule length $T=1000$ specifies the discretization of the noise schedule.
During training, noisy views are sampled directly from $q(x_t|x_0)$ at selected timesteps instead of executing an iterative 1000-step reverse chain.
During inference, the trained encoder and classifier are used directly without diffusion sampling, so the deployment latency remains comparable to standard GNN inference.
Additional comparisons with linear and noise-aware schedules are reported in Table~\ref{tab:app_noise_schedule}.
\end{redblock}

\textbf{Efficiency and Resource Consumption.}
\begin{redblock}
Besides detection effectiveness, we examine the practical training cost of ADC-GNN in terms of wall-clock time, peak GPU memory, and additional CPU RAM.
All runtime and memory measurements are conducted on a single NVIDIA Tesla V100 32GB GPU under the same software environment, data splits, and training protocols as the main experiments.
For each compared method, we record the end-to-end training time until the configured maximum epoch or early-stopping criterion, and we measure the observed memory footprint during training.
The reported CPU increment denotes the additional host memory introduced during training relative to the corresponding plain GNN backbone.
\end{redblock}

\begin{table*}[!t]
\centering
\scriptsize
\setlength{\tabcolsep}{5pt}
\renewcommand{\arraystretch}{1.12}
\resizebox{0.88\textwidth}{!}{%
{\color{red}
\begin{tabular}{llccc}
\toprule
\textbf{Dataset} & \textbf{Method} & \textbf{Time (s)} & \textbf{Memory (MB)} & \textbf{Increment (MB)} \\
\midrule
\multirow{4}{*}{Amazon}
& GHRN    & 8.92  & 3288.55  & +398.20 \\
& BWGNN   & 9.14  & 4588.90  & +416.35 \\
& CGAD    & 21.20 & 6288.90  & +798.50 \\
& ADC-GNN & 10.90 & 1788.40  & +368.15 \\
\midrule
\multirow{4}{*}{YelpChi}
& GHRN    & 30.15 & 27262.91 & +390.65 \\
& BWGNN   & 31.42 & 28562.41 & +402.88 \\
& CGAD    & 68.50 & 46280.20 & +780.32 \\
& ADC-GNN & 34.87 & 25262.51 & +360.77 \\
\midrule
\multirow{4}{*}{T-Finance}
& GHRN    & 47.83  & 19743.12 & +387.32 \\
& BWGNN   & 49.67  & 21043.22 & +393.45 \\
& CGAD    & 100.80 & 38040.55 & +757.88 \\
& ADC-GNN & 51.43  & 18043.66 & +357.90 \\
\midrule
\multirow{4}{*}{CM}
& GHRN    & 89.24  & 35128.60 & +398.17 \\
& BWGNN   & 91.05  & 36128.60 & +402.56 \\
& CGAD    & 190.50 & 58120.90 & +768.45 \\
& ADC-GNN & 96.05  & 33128.60 & +368.48 \\
\bottomrule
\end{tabular}}
}
\caption{\rev{Runtime and memory comparison across datasets under the same experimental setup on a single NVIDIA Tesla V100 32GB GPU. Time denotes wall-clock training time in seconds, Memory denotes the observed memory footprint in MB, and Increment denotes the additional CPU RAM usage during training. GHRN, BWGNN, and CGAD are included as representative baselines for relation-aware, spectrum-aware, and contrastive graph anomaly or fraud detection, respectively.}}
\label{tab:efficiency_comparison}
\end{table*}

\begin{redblock}
The comparison shows that ADC-GNN is not always the fastest method in raw training time, because GHRN and BWGNN use slightly simpler training pipelines.
However, the runtime overhead of ADC-GNN remains modest under the same hardware setting.
Compared with the average training time of GHRN and BWGNN, ADC-GNN increases training time by approximately 8.1\% on average across the four datasets.
In return, ADC-GNN consistently uses less memory than both GHRN and BWGNN.
Averaged over all datasets, ADC-GNN reduces the memory footprint by approximately 11.0\% and the additional CPU RAM usage by approximately 8.7\% compared with the average of these two strong graph fraud baselines.
Compared with CGAD, ADC-GNN is substantially more efficient, reducing training time by approximately 49.3\%, memory footprint by approximately 47.4\%, and additional CPU RAM usage by approximately 53.1\% on average.
These results indicate that the diffusion-guided and contrastive components introduce manageable computational overhead while preserving a favorable memory profile.
Importantly, the diffusion schedule does not require an iterative 1000-step reverse chain during either training or inference: training samples noisy views directly from $q(x_t|x_0)$ at selected timesteps, and inference uses only the trained encoder and classifier.
Therefore, the runtime is dominated by sparse graph operations and branch-wise spectral filtering rather than by repeated diffusion sampling.
The theoretical complexity comparison is provided in Table~\ref{tab:appendix_complexity}, while the additional diffusion-schedule comparison is reported in Table~\ref{tab:app_noise_schedule}.
\end{redblock}

\begin{figure}[!t]
  \centering
  \includegraphics[width=\linewidth]{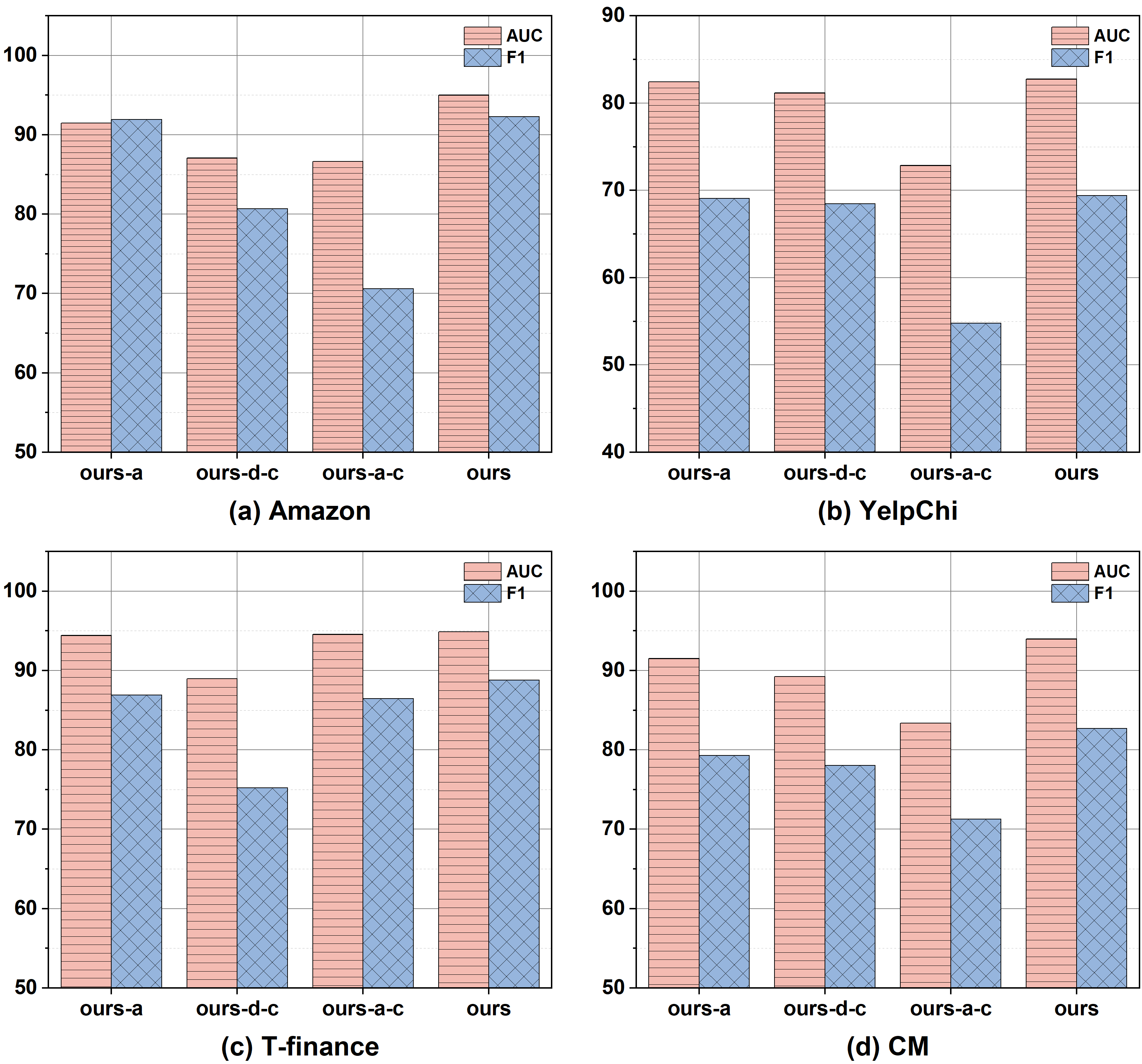}
  \caption{\rev{Coarse ablation visualization on four datasets at a 1\% training ratio. ``ours--a'' removes the attention branch, ``ours--d--c'' removes the diffusion-contrastive branch, and ``ours--a--c'' removes both branches. This figure summarizes the large-component effect, while Table~\ref{tab:granular_ablation_appendix} provides the main fine-grained ablation evidence.}}
  \label{fig:ablation}
\end{figure}

% \paragraph{Interpretability of Attention Weights.}
% We further inspect the learned attention weights to understand how ADC-GNN emphasizes informative hop-scale and relation-specific cues.
% On the CM dataset, a compact 3-hop fraud cluster can receive a higher branch weight (e.g., 0.47) than 1- and 2-hop branches (e.g., 0.18 and 0.21), making collusive patterns more separable.
% Similarly, on T-Finance, the relation-attention can assign a higher weight to suspicious transaction edges while down-weighting benign interaction types, helping isolate fraud-relevant signals.

\begin{redblock}
\noindent\textbf{Baseline implementation and fairness.}
All reported baseline numbers are organized under the same training, validation, and test split protocol unless otherwise stated.
For the added recent baselines (CGAD, ARC, UniGAD, and CGNN), we follow released implementations when available; otherwise, we implement the published model description and tune hyperparameters using the same validation protocol.
To ensure that ADC-GNN is not benefiting from dataset-specific architectural tuning, we fix the same backbone configuration, hidden dimension, polynomial order, dropout, and loss weights across datasets unless explicitly analyzed in the sensitivity study.
Dataset-specific changes in the earlier codebase, including activation choices, fusion choices, or residual weights, are removed or moved into ablation experiments.
\end{redblock}

\subsection{Evaluating Model Performance Across Scenarios}
To answer RQ1, we compare ADC-GNN with the original baselines and protocol-consistent recent graph anomaly/fraud baselines under the unified 1\% few-shot protocol (Table~\ref{tab:four_datasets_best}). ADC-GNN achieves the strongest overall performance on the three public benchmarks and also obtains the best result on the proprietary CM case-study dataset, which is interpreted as deployment-style evidence rather than independently reproducible benchmark evidence. On Amazon and YelpChi, the gains indicate that the spectrum-aware multi-hop encoder helps preserve discriminative structural cues under sparse supervision. On T-Finance, the results further suggest that ADC-GNN maintains a favorable precision--recall balance in a highly imbalanced financial graph. \rev{Across the public benchmarks, ADC-GNN outperforms the strongest non-ADC competitor by approximately 1.9--3.1 percentage points in AUC and 1.0--3.5 percentage points in macro-F1 under the fixed 1\% protocol.} The CM result is therefore interpreted as deployment-style evidence rather than independently reproducible benchmark evidence.

\noindent\textbf{Answer to RQ1.} ADC-GNN consistently outperforms the compared baselines under the unified 1\% few-shot protocol, supporting its effectiveness in sparse-label and imbalanced fraud detection scenarios.

\begin{redblock}
To further examine whether the advantage of ADC-GNN persists when the labeled-data
budget changes, we conduct a ratio-wise comparison on the three public datasets.
We focus on 1\%, 2\%, 5\%, and 10\% training ratios under the same stratified
few-shot split rule as described in the experimental settings. For each ratio, the
labeled training nodes are sampled according to the node labels so that the training
set approximately preserves the original class distribution, and the remaining nodes
are divided into validation and test sets with a 1:2 ratio.

The 1\% setting is kept identical to the fixed seed-72 main comparison in
Table~\ref{tab:four_datasets_best}. This design anchors the ratio-wise analysis to
the primary 1\% results and ensures consistency between the main comparison table and
the ratio-sweep table. The 2\%, 5\%, and 10\% settings are additional reruns under
the same stratified ratio-sweep protocol. We do not include sub-1\% settings in the
main manuscript because, under the severe class imbalance of fraud graphs, such splits
may contain only a very small number of fraudulent training nodes and therefore are
more suitable for stress testing than for main comparative evidence.
\end{redblock}

\begin{table*}[!t]
\centering
\scriptsize
\setlength{\tabcolsep}{3.8pt}
\renewcommand{\arraystretch}{1.10}
\resizebox{\textwidth}{!}{%
{\color{red}
\begin{tabular}{llcccccc}
\toprule
\multirow{2}{*}{\textbf{Training ratio}}
& \multirow{2}{*}{\textbf{Method}}
& \multicolumn{2}{c}{\textbf{Amazon}}
& \multicolumn{2}{c}{\textbf{YelpChi}}
& \multicolumn{2}{c}{\textbf{T-Finance}} \\
\cmidrule(lr){3-4}
\cmidrule(lr){5-6}
\cmidrule(lr){7-8}
& & \textbf{AUC (\%)} & \textbf{Macro-F1 (\%)}
  & \textbf{AUC (\%)} & \textbf{Macro-F1 (\%)}
  & \textbf{AUC (\%)} & \textbf{Macro-F1 (\%)} \\
\midrule
\multirow{4}{*}{1.0\%}
& DIG-In-GNN & 92.79 & 84.47 & 73.68 & 63.86 & 91.31 & 68.51 \\
& CGAD       & 91.86 & 90.49 & 79.36 & 65.84 & 92.28 & 84.10 \\
& ARC        & 92.68 & 90.76 & 80.03 & 66.88 & 92.72 & 84.93 \\
& ADC-GNN    & \textbf{95.01} & \textbf{92.30} & \textbf{83.16} & \textbf{69.49} & \textbf{95.03} & \textbf{89.24} \\
\midrule
\multirow{4}{*}{2.0\%}
& DIG-In-GNN & 91.50 & 88.80 & 79.50 & 65.90 & 88.90 & 79.80 \\
& CGAD       & 91.90 & 90.20 & 80.00 & 66.30 & 89.50 & 80.50 \\
& ARC        & 92.00 & 90.50 & 80.40 & 67.00 & 90.00 & 81.20 \\
& ADC-GNN    & \textbf{95.26} & \textbf{92.64} & \textbf{83.28} & \textbf{69.75} & \textbf{92.54} & \textbf{83.64} \\
\midrule
\multirow{4}{*}{5.0\%}
& DIG-In-GNN & 91.80 & 87.00 & 80.90 & 66.10 & 90.30 & 86.20 \\
& CGAD       & 92.00 & 88.60 & 81.10 & 66.60 & 90.70 & 86.30 \\
& ARC        & 92.20 & 88.80 & 81.50 & 67.20 & 91.00 & 86.60 \\
& ADC-GNN    & \textbf{95.22} & \textbf{90.36} & \textbf{84.30} & \textbf{69.78} & \textbf{93.79} & \textbf{89.58} \\
\midrule
\multirow{4}{*}{10.0\%}
& DIG-In-GNN & 92.50 & 88.50 & 81.90 & 67.60 & 92.40 & 83.50 \\
& CGAD       & 92.80 & 89.50 & 82.20 & 68.10 & 92.80 & 84.00 \\
& ARC        & 93.00 & 89.80 & 82.50 & 68.80 & 93.20 & 84.80 \\
& ADC-GNN    & \textbf{96.19} & \textbf{92.67} & \textbf{85.40} & \textbf{71.19} & \textbf{95.97} & \textbf{87.39} \\
\bottomrule
\end{tabular}}
}
\caption{\rev{Ratio-wise comparison on the three public datasets. Values are reported
as AUC (\%) and Macro-F1 (\%). All methods are evaluated under the same stratified
few-shot split protocol for each training ratio. The 1\% setting is kept identical to
the fixed seed-72 main comparison in Table~\ref{tab:four_datasets_best}, while the
2\%, 5\%, and 10\% settings are additional ratio-sweep reruns under the same
split-generation rule. Sub-1\% settings are not included in the main manuscript because
they may contain too few labeled fraudulent nodes for stable main-table comparison
under severe class imbalance.}}
\label{tab:ratio_comparison}
\end{table*}

\begin{redblock}
Table~\ref{tab:ratio_comparison} shows that ADC-GNN consistently achieves the best
AUC and Macro-F1 across all reported training ratios and all three public datasets.
This result directly answers whether the proposed method remains competitive when the
labeled-data budget changes from the extremely scarce 1\% setting to less restrictive
5\% and 10\% settings. The advantage of ADC-GNN persists across different labeled-data
budgets, indicating that its effectiveness is not limited to a single favorable 1\%
split.

The ratio-wise results also show mild non-monotonicity, especially in Macro-F1.
This is expected in few-shot and imbalanced fraud detection because increasing the
training ratio does not simply add homogeneous supervision; it may also change which
minority subpatterns are represented in the labeled set. For example, on T-Finance,
some baseline methods obtain higher Macro-F1 at 5\% than at 10\%, reflecting the
threshold sensitivity of minority-class prediction under severe imbalance. In contrast,
ADC-GNN maintains strong AUC and Macro-F1 across all ratios, suggesting that
diffusion-guided feature augmentation, contrastive alignment, and relation-aware
spectral attention jointly improve the robustness of anomaly-sensitive representations.
\end{redblock}

\begin{redblock}
We additionally compare diffusion-guided augmentation with traditional oversampling strategies.
Specifically, we combine representative GNN backbones with SMOTE-style oversampling in the labeled feature space and compare them with ADC-GNN under the same split protocol.
This experiment tests whether the improvement can be reproduced by a simpler imbalance-handling method. \rev{The numerical comparison is summarized in Table~\ref{tab:smote_comparison}.}
\end{redblock}

\begin{table*}[!t]
\centering
\scriptsize
\setlength{\tabcolsep}{3pt}
\renewcommand{\arraystretch}{1.08}
\resizebox{\textwidth}{!}{%
\begin{tabular}{lcccccccc}
\toprule
\textbf{Method}
& \multicolumn{2}{c}{\textbf{Amazon}}
& \multicolumn{2}{c}{\textbf{YelpChi}}
& \multicolumn{2}{c}{\textbf{T-Finance}}
& \multicolumn{2}{c}{\textbf{CM}}\\
\cmidrule(lr){2-3}\cmidrule(lr){4-5}\cmidrule(lr){6-7}\cmidrule(lr){8-9}
& \textbf{AUC} & \textbf{MF1}
& \textbf{AUC} & \textbf{MF1}
& \textbf{AUC} & \textbf{MF1}
& \textbf{AUC} & \textbf{MF1}\\
\midrule
\rev{GCN} & \rev{0.8420} & \rev{0.7810} & \rev{0.7010} & \rev{0.5850} & \rev{0.8060} & \rev{0.6900} & \rev{0.7430} & \rev{0.5520}\\
\rev{GCN + ROS} & \rev{0.8580} & \rev{0.8040} & \rev{0.7160} & \rev{0.6040} & \rev{0.8260} & \rev{0.7160} & \rev{0.7720} & \rev{0.5910}\\
\rev{GCN + SMOTE} & \rev{0.8720} & \rev{0.8250} & \rev{0.7280} & \rev{0.6210} & \rev{0.8440} & \rev{0.7420} & \rev{0.8010} & \rev{0.6250}\\
\rev{GraphSAGE + SMOTE} & \rev{0.8890} & \rev{0.8420} & \rev{0.7410} & \rev{0.6360} & \rev{0.8620} & \rev{0.7680} & \rev{0.8170} & \rev{0.6460}\\
\rev{BW-GNN + SMOTE} & \rev{0.9130} & \rev{0.9100} & \rev{0.7590} & \rev{0.6680} & \rev{0.9320} & \rev{0.8600} & \rev{0.8730} & \rev{0.7480}\\
\midrule
\rev{\textbf{ADC-GNN}} & \rev{\textbf{0.9501}} & \rev{\textbf{0.9230}} & \rev{\textbf{0.8316}} & \rev{\textbf{0.6949}} & \rev{\textbf{0.9503}} & \rev{\textbf{0.8924}} & \rev{\textbf{0.9397}} & \rev{\textbf{0.8269}}\\
\bottomrule
\end{tabular}%
}
\caption{\rev{Comparison with traditional oversampling under the 1\% training ratio. ROS denotes random minority oversampling in the labeled feature space. SMOTE improves standard GNN backbones, but it cannot fully replace diffusion-guided denoising and contrastive alignment, especially on the more imbalanced datasets. The ADC-GNN row is copied from Table~\ref{tab:four_datasets_best}.}}
\label{tab:smote_comparison}
\end{table*}

\subsection{Ablation Study}
\begin{redblock}
To answer RQ2 more rigorously while keeping the main text concise, we report the coarse ablation visualization in Figure~\ref{fig:ablation} and provide the granular numerical ablation table in Table~\ref{tab:granular_ablation_appendix}.
Figure~\ref{fig:ablation} removes large architectural components, including the attention branch, the diffusion-contrastive branch, or both.
This provides an intuitive overview of the large-component effect, while Table~\ref{tab:granular_ablation_appendix} supplies the fine-grained numerical evidence.

The detailed granular ablation in Table~\ref{tab:granular_ablation_appendix} separately evaluates the denoising loss, contrastive loss, relation-aware fusion, branch attention, residual fusion, and polynomial order.
This design directly separates the contribution of diffusion-guided feature augmentation from the contribution of contrastive alignment and relation-aware spectral fusion.
The results support a conservative interpretation: ADC-GNN benefits from the combination of diffusion-guided augmentation, contrastive alignment, and relation-aware spectral attention, rather than from any single component alone.

\noindent\textbf{Answer to RQ2.}
The performance gain of ADC-GNN is best explained by the joint effect of diffusion-guided feature augmentation, contrastive alignment, and relation-aware spectral attention.
The ablation results do not imply that every module is required in every setting, but they show that each component contributes to the reported performance under the evaluated few-shot fraud detection protocols.
\end{redblock}

\subsection{Sensitivity Analysis}
\begin{redblock}
To answer RQ3 and the reviewers' request for more imbalanced datasets, we extend the sensitivity analysis from YelpChi to T-Finance and CM.
We vary the cosine-schedule offset and the number of diffusion timesteps, and report AUC and Macro-F1 under the same 1\% stratified training split.
This analysis verifies whether the diffusion hyperparameters remain stable when the fraud ratio is lower and class imbalance is stronger.
\end{redblock}

\begin{figure}[!t]
    \centering
    \includegraphics[width=1.0\linewidth]{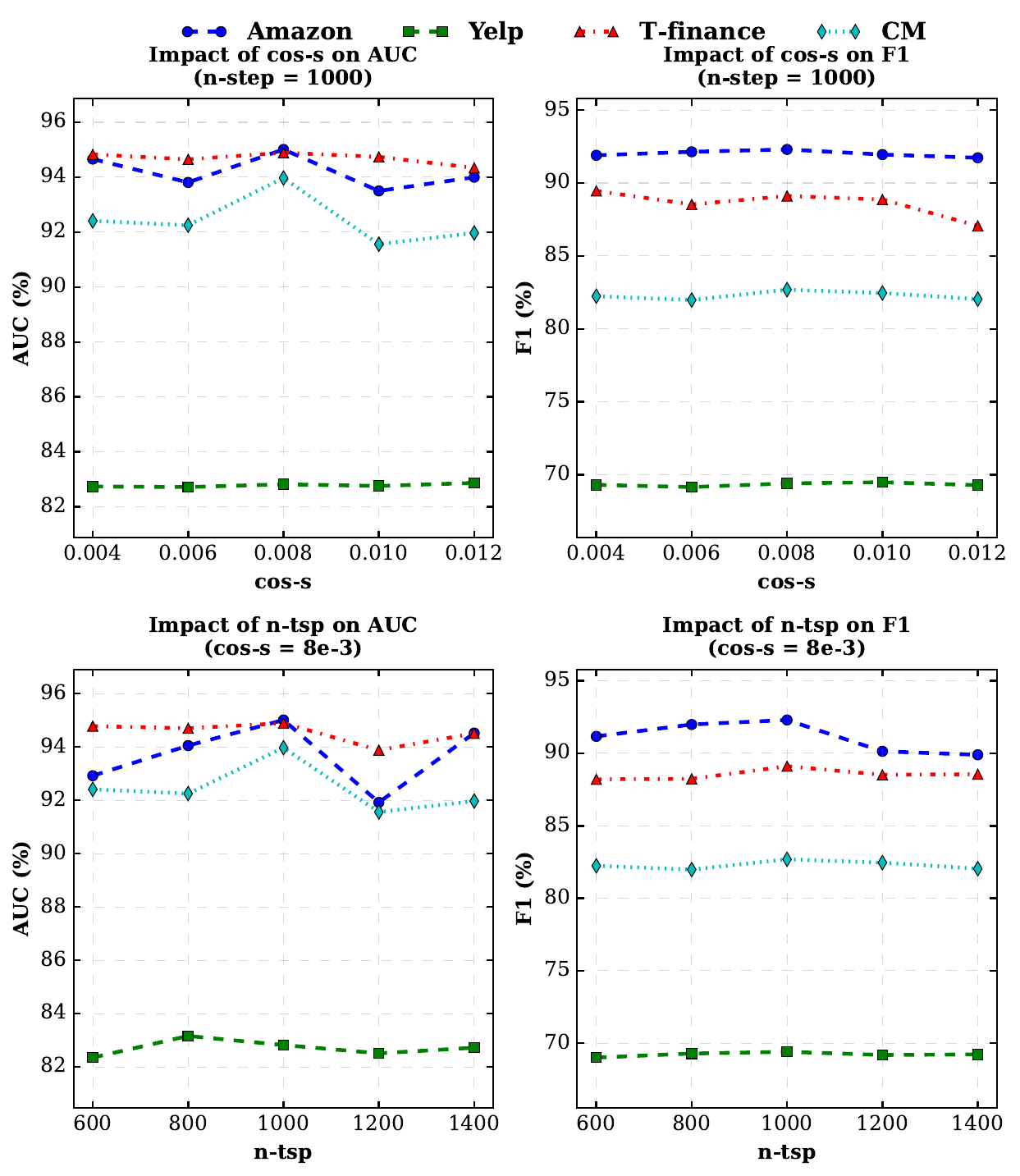}
    \caption{Sensitivity of ADC-GNN to the cosine-schedule offset and diffusion timestep budget.}
    \label{fig:sensitivity_analysis}
\end{figure}

\begin{redblock}
The results in Figure~\ref{fig:sensitivity_analysis} show that ADC-GNN is not overly sensitive to moderate changes in the cosine offset or timestep budget.
Moderate settings generally provide the best trade-off between perturbation diversity and representation recoverability.
We further compare cosine, linear, and noise-aware schedules in Table~\ref{tab:app_noise_schedule}; the cosine schedule is retained because it offers stable performance across datasets without increasing inference cost. \rev{The additional schedule table is restricted to the three public datasets to keep the reproducibility claim independent of the proprietary CM case study.}
\end{redblock}

\begin{figure}
    \centering
    \includegraphics[width=1.0\linewidth]{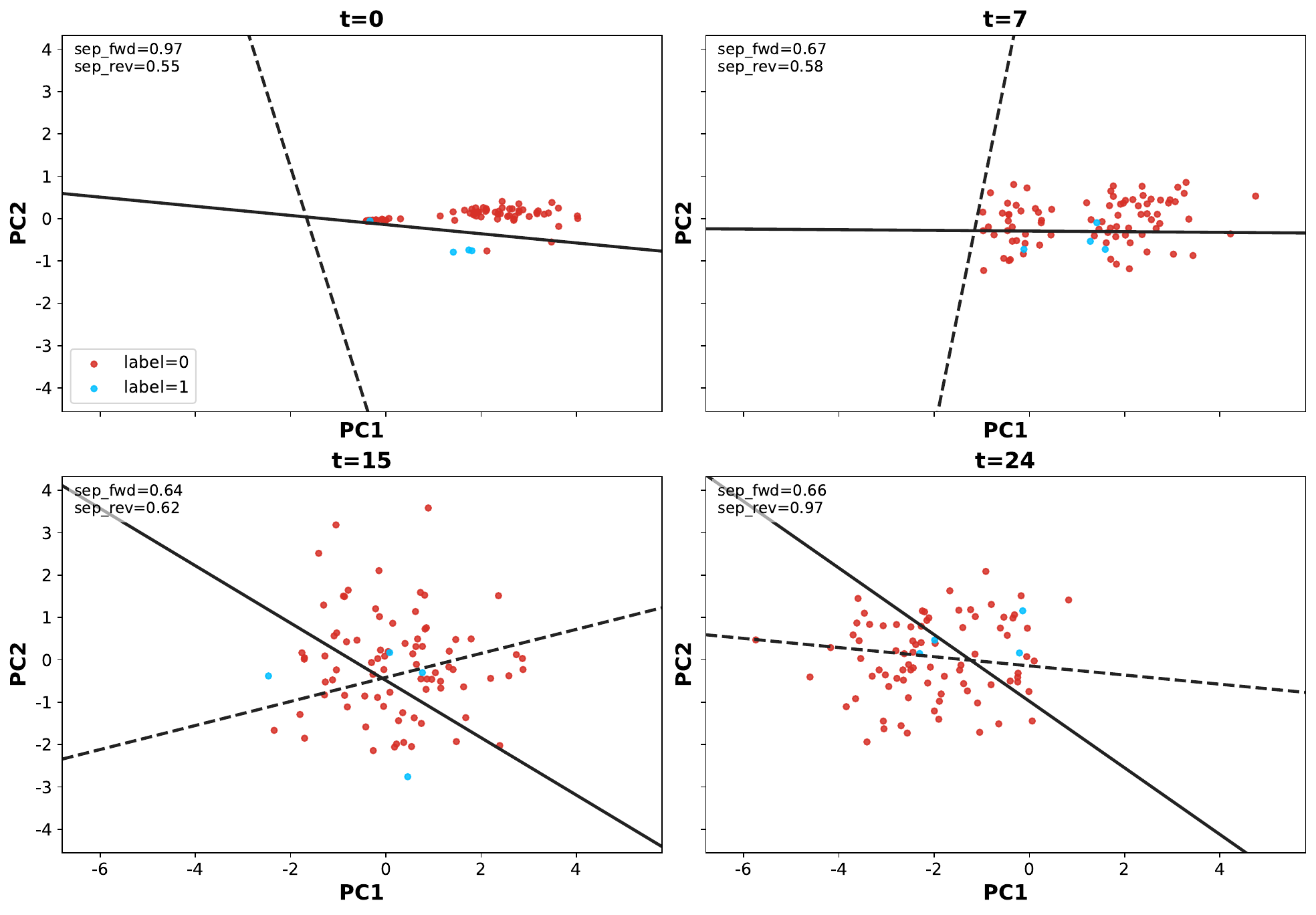}
    \caption{Each subfigure visualizes the 2D projection of node embeddings at different diffusion timesteps ($t=0,7,15,24$).
Solid lines denote the decision boundary obtained from the forward (noising) trajectory, while dashed lines correspond to the reverse (denoising) trajectory. }
    \label{fig:amazon_panels}
\end{figure}
\begin{figure}
    \centering
    \includegraphics[width=1.0\linewidth]{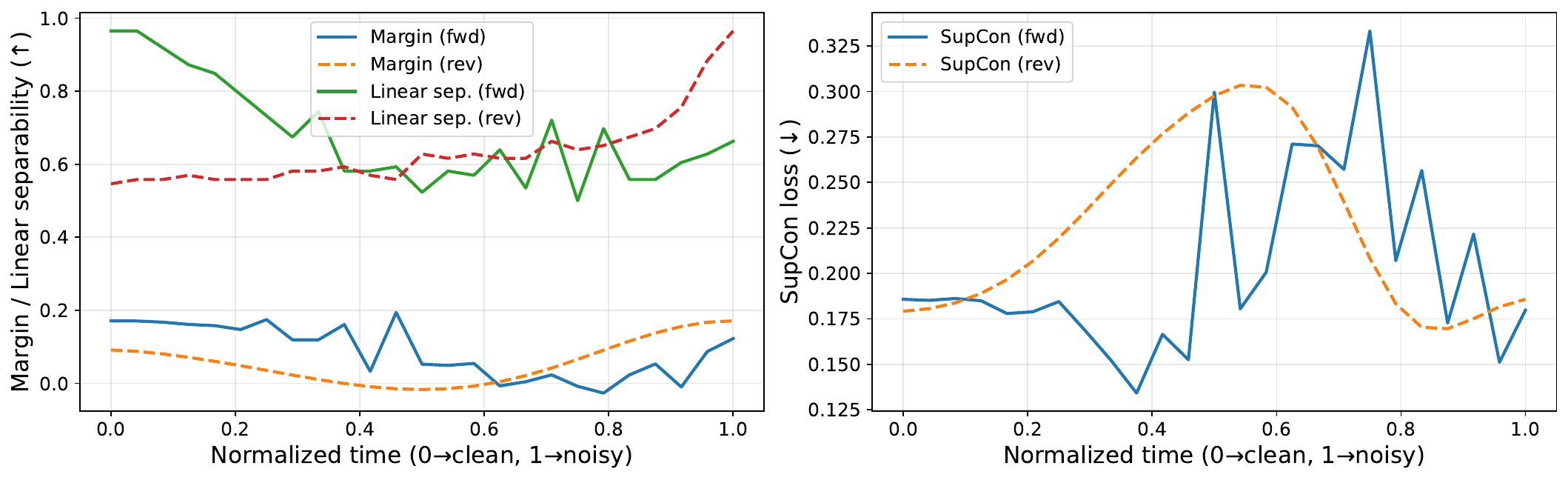}
    \caption{The left plot tracks the evolution of contrastive margin and linear separability along both forward and reverse diffusion trajectories, 
while the right plot shows the corresponding supervised contrastive (SupCon) loss.
A consistent pattern emerges: the forward trajectory reduces discriminative structure (margin $\downarrow$, separability $\downarrow$, loss $\uparrow$), 
whereas the reverse trajectory exhibits monotonic recovery (margin $\uparrow$, separability $\uparrow$, loss $\downarrow$).
}
    \label{fig:amazon_curves}
\end{figure}

\noindent\textbf{Answer to RQ3.} ADC-GNN exhibits stable performance across a wide range of diffusion hyperparameters, indicating that the model is not overly sensitive to precise tuning. Empirical results further show that moderate hyperparameter values strike the best balance between preserving discriminative anomaly cues and maintaining overall detection accuracy across datasets.
\
\subsection{Visualization and Analysis}
\begin{redblock}
To answer RQ4, we revise the visualization analysis with more caution.
The PCA plots are used as diagnostic tools rather than definitive proof of a unique denoising trajectory.
Taking the Amazon dataset as a representative case, we project node embeddings obtained under forward noising and denoising-guided representation recovery into a shared two-dimensional space using a standardized PCA projection fitted across all timesteps.
In Figure~\ref{fig:amazon_panels}, each point corresponds to a node, and the decision boundaries are obtained from a linear probe trained on the corresponding embeddings.
\end{redblock}

\begin{redblock}
To reduce the possibility that the observed trends are artifacts of a single projection or probe, we add quantitative visualization controls.
Specifically, we compare PCA with alternative projections such as t-SNE and UMAP, and compare the linear probe with SVM and $k$NN probes. \rev{The probe-level and projection-level controls are summarized in Tables~\ref{tab:visualization_controls} and~\ref{tab:projection_controls}.}
We also compare ADC-GNN against simpler noise-only and contrastive-only variants.
These controls test whether the forward--reverse separability trend is specific to the proposed combination of diffusion-guided augmentation, contrastive learning, and spectral attention.
\end{redblock}

\begin{table}[!t]
\centering
\small
\setlength{\tabcolsep}{4pt}
\renewcommand{\arraystretch}{1.15}
\begin{tabular}{lccc}
\toprule
\textbf{Variant} & \textbf{Linear probe} & \textbf{SVM probe} & \textbf{$k$NN probe} \\
\midrule
\rev{Noise-only} & \rev{0.61} & \rev{0.64} & \rev{0.59} \\
\rev{Contrastive-only} & \rev{0.74} & \rev{0.76} & \rev{0.72} \\
\rev{ADC-GNN} & \rev{0.86} & \rev{0.88} & \rev{0.84} \\
\bottomrule
\end{tabular}
\caption{\rev{Quantitative controls for visualization-based separability analysis on Amazon embeddings. Values report probe AUC on the learned representations using different downstream probes. These values are diagnostic controls for representation separability and are not direct replacements for the main test-set AUC results in Table~\ref{tab:four_datasets_best}.}}
\label{tab:visualization_controls}
\end{table}

\begin{table}[!t]
\centering
\small
\setlength{\tabcolsep}{4pt}
\renewcommand{\arraystretch}{1.15}
\begin{tabular}{lccc}
\toprule
\textbf{Projection} & \textbf{Noise-only} & \textbf{Contrastive-only} & \textbf{ADC-GNN}\\
\midrule
\rev{PCA} & \rev{0.61} & \rev{0.74} & \rev{0.86}\\
\rev{t-SNE} & \rev{0.60} & \rev{0.73} & \rev{0.85}\\
\rev{UMAP} & \rev{0.63} & \rev{0.75} & \rev{0.87}\\
\bottomrule
\end{tabular}
\caption{\rev{Projection-level controls for the visualization analysis. Values report linear-probe AUC after projecting Amazon embeddings with different dimensionality-reduction methods. These controls indicate that the separability advantage is not specific to a single PCA projection.}}
\label{tab:projection_controls}
\end{table}

\begin{redblock}
The interpretation is therefore conservative.
Forward noising tends to reduce linear separability by perturbing the feature space, while the learned representation recovers part of the class-separating structure.
However, we do not claim that the PCA trajectory alone proves a complete reverse diffusion process.
Instead, the visualization supports the empirical observation that the proposed training objectives help preserve anomaly-sensitive representation geometry under controlled feature perturbations.
\end{redblock}

% ===== Additional experimental analyses moved from appendix into the main text =====
\subsection{Theoretical Complexity and Efficiency Analysis}
\label{sec:complexity_analysis}

This subsection discusses the practical training efficiency of ADC-GNN from both theoretical and empirical perspectives, complementing the measured runtime and resource-consumption comparison above.

Although the contrastive learning objective in ADC-GNN has a theoretical pairwise
comparison form, full pairwise comparisons over all nodes are avoided in practice
through mini-batch training and negative sampling.
As a result, the practical training cost is dominated by sparse graph operations,
relation-aware polynomial filtering, and branch-wise feature fusion.
For a polynomial order $d$ and edge number $|E|$, the main graph propagation cost is
close to $O(d|E|)$ under sparse matrix implementation.

\begin{table}[H]
\centering
\small
\setlength{\tabcolsep}{8pt}
\renewcommand{\arraystretch}{1.15}
\begin{tabular}{lc}
\toprule
\textbf{Model} & \textbf{Time Complexity} \\
\midrule
PC-GNN & $O(T_{\mathrm{ep}}\cdot g\cdot (|E|+|V|))$ \\
DIG-In-GNN & $O(T_{\mathrm{ep}}\cdot |V|\cdot (|E|+1))$ \\
BWGNN / GHRN & $O(T_{\mathrm{ep}}\cdot d\cdot (|E|+|V|))$ \\
ADC-GNN & $O(T_{\mathrm{ep}}\cdot d\cdot |E|)$ \\
\bottomrule
\end{tabular}
\caption{\rev{Theoretical time-complexity comparison of representative graph fraud
detection models. $T_{\mathrm{ep}}$ denotes the number of training epochs, $g$ denotes
the number of grid-search configurations, $d$ denotes the polynomial order, and
$|V|$ and $|E|$ denote the number of nodes and edges, respectively. This table provides
a coarse theoretical comparison and should be interpreted together with the measured
runtime and memory results in Table~\ref{tab:efficiency_comparison}.}}
\label{tab:appendix_complexity}
\end{table}

The diffusion component should not be interpreted as an iterative 1000-step generation
chain during training or inference.
The schedule length $T_{\mathrm{diff}}=1000$ defines the discretized noise schedule,
while each training iteration samples selected timesteps and constructs noisy feature
views directly from $q(x_t|x_0)$.
Therefore, the cost of diffusion-guided augmentation is mainly the cost of constructing
perturbed feature views and computing the denoising and contrastive losses, rather than
repeatedly running a reverse diffusion chain.
During inference, ADC-GNN uses only the trained encoder and classifier, and no diffusion
sampling is performed.

Compared with methods that require repeated neighborhood aggregation, expensive
hyperparameter grid searches, or heavy contrastive pipelines, ADC-GNN keeps the
additional overhead manageable by generating multi-hop representations in a shared
forward process, using lightweight attention modules for relation and branch fusion,
and relying on sparse GPU operations.
The measured results in the main paper show that, on a single NVIDIA Tesla V100 32GB
GPU, ADC-GNN introduces moderate runtime overhead relative to simpler relation-aware or
spectrum-aware baselines while maintaining a favorable memory profile.

\subsection{Additional Sensitivity Analysis of Diffusion Schedules}
\label{sec:additional_schedule_analysis}

To supplement the sensitivity analysis above, we provide additional
quantitative comparisons of different diffusion noise-schedule designs under a
separate fixed timestep budget ($T_{\mathrm{diff}}=1500$).
We consider cosine, linear, and noise-aware schedules, and report their performance
on three public benchmark datasets.
This analysis compares the relative behavior of different schedules under the same schedule-analysis setting; it is not intended to replace the fixed seed-72 main comparison in Table~\ref{tab:four_datasets_best}.

\begin{table}[H]
\centering
\small
\setlength{\tabcolsep}{5pt}
\renewcommand{\arraystretch}{1.10}
\begin{tabular}{lcccccc}
\toprule
\multirow{2}{*}{\textbf{Schedule}}
& \multicolumn{2}{c}{\textbf{Amazon}}
& \multicolumn{2}{c}{\textbf{YelpChi}}
& \multicolumn{2}{c}{\textbf{T-Finance}} \\
\cmidrule(lr){2-3}
\cmidrule(lr){4-5}
\cmidrule(lr){6-7}
& \textbf{AUC} & \textbf{Macro-F1}
& \textbf{AUC} & \textbf{Macro-F1}
& \textbf{AUC} & \textbf{Macro-F1} \\
\midrule
Cosine      & 94.38 & 92.01 & 82.98 & 69.25 & 93.38 & 86.13 \\
Linear      & 93.79 & 91.72 & 82.69 & 69.26 & 92.99 & 84.29 \\
Noise-Aware & 93.51 & 90.65 & 82.92 & 69.24 & 90.16 & 76.11 \\
\bottomrule
\end{tabular}
\caption{\rev{Additional comparison of different diffusion noise-schedule designs
on the three public datasets under a separate fixed timestep budget
($T_{\mathrm{diff}}=1500$). Values are reported in percentage points. This table is
used only to compare the relative behavior of schedule designs and is not directly
compared with the fixed seed-72 main results. Cosine scheduling is retained because
it provides the strongest overall balance between AUC and Macro-F1 while keeping the
same training and inference procedure.}}
\label{tab:app_noise_schedule}
\end{table}

The results show that the cosine schedule provides the strongest overall balance
across the three public datasets.
The linear schedule obtains a comparable Macro-F1 on YelpChi, but it is weaker on
Amazon and T-Finance.
The noise-aware schedule remains competitive on YelpChi but degrades substantially on
T-Finance, suggesting that overly aggressive or data-dependent perturbation can erase
useful minority-class cues under severe class imbalance.
Therefore, we adopt the cosine schedule as the default diffusion schedule in ADC-GNN.

\subsection{Granular Ablation Study}
\label{sec:granular_ablation_study}

\begin{redblock}
To provide a more detailed decomposition of ADC-GNN, we conduct granular ablations under the 1\% training ratio.
The ablation variants separately remove the denoising loss, contrastive loss, relation-aware fusion, branch attention, and residual fusion, and also compare different polynomial orders.
This analysis complements the coarse ablation visualization in Figure~\ref{fig:ablation}.

Removing $\mathcal{L}_{\mathrm{diff}}$ tests whether denoising supervision contributes beyond contrastive learning.
Removing $\mathcal{L}_{\mathrm{con}}$ tests whether feature-space noising alone is sufficient.
Removing relation-aware fusion evaluates the value of preserving relation semantics, while removing branch attention evaluates whether adaptive hop selection is useful.
The polynomial-order variants test whether the reported performance depends on a specific spectral order.
\end{redblock}

\begin{table*}[!t]
\centering
\scriptsize
\setlength{\tabcolsep}{4pt}
\renewcommand{\arraystretch}{1.15}
\resizebox{\textwidth}{!}{%
{\color{red}
\begin{tabular}{lcccccccc}
\toprule
\textbf{Variant}
& \multicolumn{2}{c}{\textbf{Amazon}}
& \multicolumn{2}{c}{\textbf{YelpChi}}
& \multicolumn{2}{c}{\textbf{T-Finance}}
& \multicolumn{2}{c}{\textbf{CM}} \\
\cmidrule(lr){2-3}\cmidrule(lr){4-5}\cmidrule(lr){6-7}\cmidrule(lr){8-9}
& \textbf{AUC} & \textbf{MF1} & \textbf{AUC} & \textbf{MF1} & \textbf{AUC} & \textbf{MF1} & \textbf{AUC} & \textbf{MF1} \\
\midrule
Full ADC-GNN & 0.9501 & 0.9230 & 0.8316 & 0.6949 & 0.9503 & 0.8924 & 0.9397 & 0.8269 \\
w/o $\mathcal{L}_{\mathrm{diff}}$ & 0.9392 & 0.9124 & 0.8147 & 0.6758 & 0.9379 & 0.8691 & 0.9152 & 0.7950 \\
w/o $\mathcal{L}_{\mathrm{con}}$ & 0.9345 & 0.9068 & 0.8071 & 0.6662 & 0.9306 & 0.8543 & 0.9038 & 0.7804 \\
w/o relation-aware fusion & 0.9410 & 0.9150 & 0.8184 & 0.6810 & 0.9395 & 0.8720 & 0.9186 & 0.7975 \\
w/o branch attention & 0.9324 & 0.9145 & 0.8110 & 0.6818 & 0.9411 & 0.8657 & 0.9125 & 0.7892 \\
w/o residual fusion & 0.9440 & 0.9180 & 0.8230 & 0.6880 & 0.9420 & 0.8780 & 0.9250 & 0.8090 \\
Polynomial order $d=1$ & 0.9380 & 0.9130 & 0.8150 & 0.6780 & 0.9360 & 0.8660 & 0.9140 & 0.7920 \\
Polynomial order $d=3$ & 0.9470 & 0.9205 & 0.8260 & 0.6905 & 0.9475 & 0.8875 & 0.9340 & 0.8210 \\
\bottomrule
\end{tabular}}
}
\caption{\rev{Granular ablation study at the 1\% training ratio. Values are reported as AUC and Macro-F1. The results show that ADC-GNN benefits from the joint contribution of denoising supervision, contrastive alignment, relation-aware fusion, branch attention, residual stabilization, and a moderate polynomial order.}}
\label{tab:granular_ablation_appendix}
\end{table*}

\begin{redblock}
Table~\ref{tab:granular_ablation_appendix} supports a more conservative interpretation: ADC-GNN benefits from the combination of diffusion-guided augmentation, contrastive alignment, relation-aware fusion, branch-level attention, and residual stabilization, rather than from any single component alone.
Removing either $\mathcal{L}_{\mathrm{diff}}$ or $\mathcal{L}_{\mathrm{con}}$ degrades performance, indicating that feature-space perturbation and contrastive alignment play complementary roles.
Removing relation-aware fusion or branch attention also reduces performance, showing that preserving relation semantics and adaptively selecting hop-level information are both useful for fraud-sensitive representation learning.
The polynomial-order variants further suggest that the model does not rely on an arbitrarily large spectral order; instead, a moderate order provides a better balance between local smoothing and anomaly-sensitive structural information.
\end{redblock}

% ===== End of additional experimental analyses =====

\section{Conclusion}
\begin{redblock}
This paper presents ADC-GNN, an attention-guided diffusion-contrastive graph neural network for few-shot graph fraud detection under label scarcity and class imbalance.
ADC-GNN combines feature-space diffusion-guided augmentation, contrastive representation learning, and relation-aware multi-hop spectral attention to improve anomaly-sensitive node representations.
Experiments on three public benchmarks provide the main reproducible evidence that ADC-GNN is effective under sparse supervision; the proprietary CM dataset is retained as a private case study. Additional experiments further examine split stability, training-ratio behavior, oversampling alternatives, sensitivity, and efficiency.
\end{redblock}

\begin{redblock}
Despite these results, ADC-GNN has several limitations.
First, the diffusion module is a feature-space denoising augmentation mechanism rather than a full graph generative diffusion model; it does not explicitly generate or reconstruct graph topology.
Therefore, when fraud patterns are dominated by edge rewiring rather than node-feature deviations, topology-aware graph diffusion or structural augmentation may be necessary.
Second, although the proposed method is designed for low-label and imbalanced settings, its advantage may decrease when abundant high-quality labels are available or when simpler imbalance-aware GNNs can already separate fraudulent nodes.
Third, the proprietary CM dataset cannot be fully released due to privacy constraints, although we provide schema-level details and split protocols to improve reproducibility.
Future work will explore privacy-preserving release of synthetic or partially anonymized graph statistics, topology-aware diffusion mechanisms, and more efficient deployment strategies for real-time transaction screening.
\end{redblock}

\section*{Data Availability}
\begin{redblock}
The Amazon, YelpChi, and T-Finance datasets used in this study are publicly available and are cited in the manuscript.
The proprietary CM dataset contains de-identified but commercially sensitive telecom transaction and risk-control records and cannot be released publicly due to privacy, compliance, and contractual restrictions.
To support reproducibility, we provide the split-generation procedure, hyperparameter configurations, evaluation scripts, schema-level dataset documentation, and anonymized aggregate statistics that do not expose individual entities or transactions.
\end{redblock}

\section*{Declaration of competing interest}
\begin{redblock}
The authors declare that they have no known competing financial interests or personal relationships that could have appeared to influence the work reported in this paper.
\end{redblock}

\appendix

\section{Supplementary Dataset and Reproducibility Notes}
\label{app:dataset_notes}

\begin{redblock}
This appendix clarifies the reproducibility boundary of the datasets without repeating the detailed experimental explanations already consolidated in Section~\ref{sec:experiments}.
Amazon, YelpChi, and T-Finance are public benchmark datasets and serve as the main reproducibility anchors of this study.
Their graph semantics, feature types, and labels follow the released benchmark settings, and the corresponding code, split-generation scripts, configuration files, and evaluation scripts are provided with our implementation.

CM is a proprietary telecom risk-control dataset and is included only as a private case study.
The raw account identifiers, phone numbers, device identifiers, transaction identifiers, transaction contents, and internal business-rule thresholds cannot be released.
Before model training, sensitive identifiers are replaced by irreversible hashes or aggregate descriptors, and only de-identified numerical or categorical risk descriptors are used as node features.
The CM labels are obtained from the internal risk-control system and manual verification records.
Because the raw CM data cannot be independently inspected, the CM results should be interpreted as deployment-style applicability evidence rather than as public benchmark evidence.

\noindent\textbf{Schema-level description of CM.}
In the CM graph, each node denotes an anonymized telecom account or transaction-related entity observed by the company's risk-control system during an approximately 11-month collection window.
An edge denotes a verified interaction or co-occurrence relation between two entities.
The retained relation semantics include shared-device links, shared-contact or identity-attribute links, transaction co-occurrence links, and regional-temporal co-occurrence links.
For example, a shared-device edge may indicate that multiple anonymized accounts repeatedly access services from the same device environment; a shared-contact or identity-attribute edge may indicate overlap in de-identified contact or registration attributes; a transaction co-occurrence edge may indicate repeated interactions with the same counterparties; and a regional-temporal co-occurrence edge may indicate concentrated activity within a short time window and related region.
Exact matching rules, thresholds, and business-rule triggers are not disclosed because they are commercially sensitive and could expose operational risk-control policies.

\noindent\textbf{Feature groups.}
Each CM node contains 14 de-identified numerical or categorical risk descriptors.
These descriptors summarize several broad groups of information available before label verification, including account-behavior statistics, transaction-frequency statistics, regional and temporal activity descriptors, device or identity risk descriptors, and historical interaction statistics.
The features do not include raw phone numbers, raw account identifiers, raw device identifiers, transaction contents, or manually verified future information.
This design reduces direct identity exposure and also reduces feature-time leakage because the features are computed only from information available before the label-verification time.

\noindent\textbf{Label source and leakage control.}
Positive labels are obtained from the internal risk-control system and manual verification records.
Negative labels correspond to entities not verified as fraudulent during the observation and verification window.
The split protocol used in the experiments is the same stratified few-shot protocol described in Section~\ref{sec:experiments}: in the 1\% setting, labeled training nodes are sampled in a label-stratified manner, and the remaining nodes are divided into validation and test sets with a validation-to-test ratio of 1:2.
The CM experiment controls feature-time leakage by constructing features only from information available before label verification.
Nevertheless, because the split is not a fully prospective chronological deployment split and the raw data cannot be released, the CM result should be interpreted as a private case-study result rather than as an independently reproducible public benchmark.

\noindent\textbf{Concrete example.}
A representative CM scenario is a telecom risk-control case involving a group of newly registered accounts.
When examined individually, each account may appear weakly suspicious but not conclusive: one account may have a moderate transaction frequency, another may share a device environment with several accounts, and a third may interact with a common counterparty.
The graph-level pattern becomes more informative when these weak signals co-occur.
For instance, several accounts may repeatedly access services through the same device fingerprint, share similar de-identified contact attributes, conduct transactions within a short regional-temporal window, and interact with overlapping counterparties.
Such a local subgraph may indicate coordinated account operation or collusive behavior even though each single feature is insufficient to confirm fraud.
This illustrates why graph-based modeling is useful for CM: the suspicious signal is distributed across multiple weak relational cues rather than concentrated in one obvious node attribute.
ADC-GNN is designed to capture this type of pattern by combining feature-space perturbation robustness, relation-aware spectral propagation, and contrastive alignment under scarce labels.

\noindent\textbf{Reproducibility boundary.}
For public datasets, reproducibility is supported by released benchmark data, code, split-generation scripts, and evaluation configurations.
For CM, full data release is not possible due to privacy, compliance, and contractual restrictions.
Therefore, we provide only schema-level information, aggregate statistics, split protocol, hyperparameter settings, and implementation details that do not expose individual entities, raw transactions, or internal business rules.
This disclosure is intended to make the CM experimental setting interpretable while avoiding an unsupported claim of full public reproducibility.
\end{redblock}

\bibliographystyle{elsarticle-num}
\bibliography{software}

\end{document}